%% file: ms.tex
\documentclass[conference,compsoc,table,xcdraw,x11names]{IEEEtran}

\usepackage[style=numeric-comp,natbib=true]{biblatex}
\input{preamble}

\newcommand\copyrighttext{%
  \footnotesize \textcopyright 2025 IEEE. Personal use of this material is permitted.
  Permission from IEEE must be obtained for all other uses, in any current or future
  media, including reprinting/republishing this material for advertising or promotional
  purposes, creating new collective works, for resale or redistribution to servers or
  lists, or reuse of any copyrighted component of this work in other works.}
\newcommand\copyrightnotice{%
\begin{tikzpicture}[remember picture,overlay]
\node[anchor=south,yshift=10pt] at (current page.south) 
  {\fbox{\parbox{\dimexpr\textwidth-\fboxsep-\fboxrule\relax}{\copyrighttext}}};
\end{tikzpicture}%
}

\usepackage{array}
\ifCLASSOPTIONcompsoc
  \usepackage[caption=false,font=footnotesize,labelfont=sf,textfont=sf]{subfig}
\else
  \usepackage[caption=false,font=footnotesize]{subfig}
\fi

\begin{document}
%
\title{Differentially Private Selection using Smooth Sensitivity}

\newcommand{\linebreakand}{%
  \end{@IEEEauthorhalign}
  \hfill\mbox{}\par
  \mbox{}\hfill\begin{@IEEEauthorhalign}
}

\author{\IEEEauthorblockN{Iago C. Chaves}
\IEEEauthorblockA{Universidade Federal do Ceará\\
Fortaleza, Ceará, Brazil\\
Email: iago.chaves@lsbd.ufc.br}
\and
\IEEEauthorblockN{Victor A. E. Farias}
\IEEEauthorblockA{Universidade Federal do Ceará\\
Fortaleza, Ceará, Brazil\\
Email: victor.farias@lsbd.ufc.br}
\and
\IEEEauthorblockN{Amanda Perez}
\IEEEauthorblockA{Fundação Getulio Vargas\\
Rio de Janeiro, Rio de Janeiro, Brazil\\
Email: perez.amanda@fgv.edu.br}
\linebreakand
\IEEEauthorblockN{Diego Mesquita}
\IEEEauthorblockA{Fundação Getulio Vargas\\
Rio de Janeiro, Rio de Janeiro, Brazil\\
Email: diego.mesquita@fgv.br}
\and
\IEEEauthorblockN{Javam C. Machado}
\IEEEauthorblockA{Universidade Federal do Ceará\\
Fortaleza, Ceará, Brazil\\
Email: javam.machado@lsbd.ufc.br}}


%
\maketitle
\copyrightnotice

\begin{abstract}
%
%
Differentially private selection mechanisms offer strong privacy guarantees for queries aiming to identify the top-scoring element $r$ from a finite set $\OUT$, based on a dataset-dependent utility function. While selection queries are fundamental in data science, few mechanisms effectively ensure their privacy. Furthermore, most approaches rely on \emph{global sensitivity} to achieve differential privacy (DP), which can introduce excessive noise and impair downstream inferences. To address this limitation, we propose the \emph{\metname} (\met) mechanism, which leverages \emph{smooth sensitivity} to yield provably tighter (upper bounds on) expected errors compared to global sensitivity-based methods. Empirical results demonstrate that \met is more accurate than state-of-the-art differentially private selection methods in three applications: percentile selection, greedy decision trees, and random forests. \looseness=-1
\end{abstract}

%
\IEEEpeerreviewmaketitle

\input{sections/1-intro}
\input{sections/2-background}
\input{sections/3-priv_selection}
\input{sections/4-method}
\input{sections/5-app_perc}
\input{sections/6-app_dt}
\input{sections/7-app_rf}
\input{sections/8-conclusion}

\ifCLASSOPTIONcompsoc
  \section*{Acknowledgments}
\else
  \section*{Acknowledgment}
\fi
This work was partially supported by CAPES/Brazil under grant number 88882.454584/2019-01. 
This work was also supported by the Laboratório de Sistemas e Banco de dados (LSBD), the Fundação Carlos Chagas Filho de Amparo à Pesquisa do Estado do Rio de Janeiro FAPERJ (SEI-260003/000709/2023), the São Paulo Research Foundation FAPESP (2023/00815-6), and the Conselho Nacional de Desenvolvimento Científico e Tecnológico CNPq (404336/2023-0).

\printbibliography

\onecolumn
\begin{appendices}
    \section{\metname proofs}
    \subsection{Privacy proof}\label{apx:priv}
    \printProofs[priv]
    \subsection{Utility proof} \label{apx:util}
    \printProofs[util]
\end{appendices}

\twocolumn

\section{Meta-Review}
The following meta-review was prepared by the program committee for the 2025
IEEE Symposium on Security and Privacy (S\&P) as part of the review process as
detailed in the call for papers.

\subsection{Summary}
This paper considers the differentially private selection problem, in which we must select an item from a set based on a dataset-dependent utility function, with differential privacy. The authors propose an algorithm called Smooth Noisy Max (SNM), which uses the notion of smooth sensitivity to reduce the error of classical algorithms (both theoretically and in practice). The authors demonstrate the utility of their approach on several downstream problems.

\subsection{Scientific Contributions}
\begin{itemize}
    \item Provides a Valuable Step Forward in an Established Field
\end{itemize}

\subsection{Reasons for Acceptance}
\begin{enumerate}
    \item The selection problem is relatively old. This paper proposes a new algorithm and theoretical analysis that outperforms widely-used methods depending on global sensitivity. The results in this work are of theoretical interest, and can be of practical interest for some problem settings.
\end{enumerate}


\end{document}

%% file: preamble.tex
\usepackage{xcolor}
\PassOptionsToPackage{table,xcdraw}{xcolor}

\usepackage{balance}

\usepackage{multicol}

\usepackage{amsmath}
\usepackage{amssymb}
\usepackage{amsthm}
\usepackage{mathtools}
\usepackage[mathscr]{eucal}
\usepackage{nicefrac}
\usepackage{calc}
\usepackage{bm}

\usepackage[hidelinks]{hyperref}
\usepackage{cleveref}

\usepackage[createShortEnv, conf={text link={See the proof in the \hyperref[sec:apx]{appendix}.}}]{proof-at-the-end}

\usepackage[inline]{enumitem}

\usepackage{multirow}
\usepackage{arydshln}

\usepackage{tikz}
\usepackage{pgfplots}
\usepackage{pgfplotstable}

\pgfplotsset{compat=newest}
\usepgfplotslibrary{groupplots}
\usetikzlibrary{calc}
\usepgfplotslibrary{fillbetween}

\usepackage[english,ruled,lined,vlined,linesnumbered]{algorithm2e}
\usepackage{algorithmic}

\DeclareMathOperator*{\argmax}{arg\,max}

\theoremstyle{definition}
\newtheorem{theorem}{Theorem}[section]
\newtheorem{lemma}[theorem]{Lemma}
\newtheorem{proposition}[theorem]{Proposition}
\newtheorem{corollary}[theorem]{Corollary}
\newtheorem{definition}[theorem]{Definition}

\SetKwRepeat{Do}{do}{while}

\newcommand{\met}{SNM\xspace}
\newcommand{\metname}{Smooth Noisy Max\xspace}

\newcommand{\Y}{\mathscr{Y}}

\newcommand{\e}{\varepsilon}
\newcommand{\univ}{\mathscr{X}}
\newcommand{\DB}{\univ}
\newcommand{\db}[1]{\mathbf{#1}}
\newcommand{\OUT}{\mathscr{R}}
\newcommand{\score}{u}
\newcommand{\nscore}{\tilde{\score}}

\newcommand{\RLEXP}{\mathscr{A}}
\newcommand{\LEXP}{\RLEXP_{u, \varepsilon}}

\newcommand{\MEXP}{\mathscr{M}^\texttt{exp}}
\newcommand{\MPF}{\mathscr{M}^\texttt{pf}}
\newcommand{\MLD}{\mathscr{M}^\texttt{dam}}
\newcommand{\RNMEXP}{\mathscr{N}^\texttt{exp}}
\newcommand{\RNMGUM}{\mathscr{N}^\texttt{gum}}

\newcommand{\SSENS}{\mathscr{S}}
\newcommand{\SSen}{\mathscr{S}_{u,\beta}}

\pgfplotsset{
    discard if not/.style 2 args={
        x filter/.code={
            \edef\tempa{\thisrow{#1}}
            \edef\tempb{#2}
            \ifx\tempa\tempb
            \else
                \def\pgfmathresult{inf}
            \fi
        }
    }
}

\newcommand{\addplotwitherrorband}[9]{
    \addplot+[name path=lower,draw=none,mark=none,y filter/.code=\pgfmathparse{\thisrow{#3} + \pgfmathresult},forget plot] table[x=#2,y=#4] {#1};
    \addplot+[name path=upper,draw=none,mark=none,y filter/.code=\pgfmathparse{\thisrow{#3} - \pgfmathresult},forget plot] table[x=#2,y=#5] {#1};
    \addplot+[fill=#6,opacity=0.0,forget plot] fill between[of=upper and lower];

    \addplot+[mark=#8, #7, thick, color=#6, mark size=1.5pt, mark options={fill=none, solid}] table [y=#3, x=#2]{#1};
        \addlegendentry{#9}
}

\newcommand{\addplotwitherrorbanddepth}[9]{
    \addplot+[name path=lower,draw=none,mark=none,y filter/.code=\pgfmathparse{\thisrow{#3} + \pgfmathresult},forget plot, discard if not={depth}{#9}] table[x=#2,y=#4] {#1};
    \addplot+[name path=upper,draw=none,mark=none,y filter/.code=\pgfmathparse{\thisrow{#3} - \pgfmathresult},forget plot, discard if not={depth}{#9}] table[x=#2,y=#5] {#1};
    \addplot+[fill=#6,opacity=0.0,forget plot] fill between[of=upper and lower];

    \addplot+[mark=#8, #7, thick, color=#6, mark size=1.5pt, mark options={fill=none, solid}, discard if not={depth}{#9}] table [y=#3, x=#2]{#1};
    \addplotwitherrorbanddepthrelay
}

\newcommand{\addplotwitherrorbanddepthrelay}[1] {
    \addlegendentry{#1}
}

\newcommand{\textover}[3][l]{%
 \makebox[\widthof{#3}][#1]{#2}%
}

%% file: sections/1-intro.tex
\section{Introduction}
\label{cap:intro}


Differential privacy (DP) establishes a mathematically rigorous framework to avoid information leakage upon releasing the outcome of a query. More formally, achieving DP entails ensuring that the outcome of a query is statistically near-indistinguishable for similar databases. This is typically done by endowing the original query with a \emph{mechanism}, which randomizes the query's output. Importantly, DP mechanisms are tailored to the space of outcomes of the query they aim to protect.


In particular, private selection mechanisms address non-numerical queries and play a crucial role in private machine learning and data analysis, with applications in classification \cite{kotsiantis2007supervised},
synthetic data generation \cite{chen2015differentially, zhang2017privbayes}, dimensionality reduction \cite{chaudhuri2013near}, and top-$k$ queries \cite{ilyas2008survey}. 
%
However, despite their vast applicability, there exist only a few mechanisms for private selection, including the exponential mechanism \cite{mcsherry2007mechanism}, the report-noisy-max algorithm \cite{dwork2014algorithmic}, permute-and-flip \cite{mckenna2020permute}, and the local dampening mechanism \cite{farias2023local}.

Most of these algorithms are based on adding noise depending on the notion of global sensitivity, which measures the most significant impact over the utility function of adding or removing an entry from all possible databases and for all possible outcomes. This approach is guided by the worst-case scenario, which usually adds high noise \cite{zhang2015private, gonem2018smooth, DBLP:journals/corr/abs-2003-00505, DBLP:conf/nips/BunS19}, potentially harming the accuracy of results. To mitigate that, \citeauthor{nissim2007smooth} \cite{nissim2007smooth} propose the concept of smooth sensitivity, an instance-based sensitivity that depends locally on the input database $\db{x}$; nevertheless, their paper focus only on numerical queries \cite{nissim2007smooth}. The local dampening algorithm already applies a similar concept (local sensitivity) to the context of selection queries \cite{farias2023local}, but it still shows some limitations, mainly related to stability and time complexity.

We propose a novel differentially private selection algorithm, termed \metname (\met), which employs smooth sensitivity for noise addition. Specifically, \met corrupts the utility score of each potential outcome \( r \in \OUT \) with random noise (e.g., from a Laplace, Laplace Log-Normal, or Student's T distribution) scaled according to a factor proportional to the instance-based sensitivity.  
Notably, we show that \met is provably more accurate than alternatives based on global sensitivity and produces better empirical results than the prior art.

\subsubsection*{Problem Statement}
We address the challenge of \textit{private data selection}, aiming to ensure that the selection process remains both privacy-preserving and capable of producing meaningful outcomes. Let \(\db{x} \in \DB\) be a sensitive database, represented as a \textit{multiset} of records from \(\DB\), where each entry \(x_i\) corresponds to a record in \(\DB\). Consider a data selection function \(f: \DB \rightarrow \OUT\) that takes \(\db{x}\) as input and produces an outcome \(r \in \OUT\). The central challenge is to release \(f(\db{x})\) in a differentially private manner—ensuring that the output reveals minimal information about any individual record in \(\db{x}\)—while maintaining the utility and relevance of the results. \looseness=-1

\subsubsection*{Contributions}
The main contribution of this paper is the first primitive for private selection relying on smooth sensitivity. Moreover, applying this concept to non-numerical selection requires addressing significant technical obstacles. For instance, the most intuitive way to adapt existing algorithms (e.g., exponential mechanism) is replacing the global sensitivity by the smooth one. However, our Theorem \ref{the:sam} shows this does not result in a differentially private algorithm. We also provide utility guarantees showing \met is never worse than existing methods under mild conditions.
In summary, the contributions of this work are:
\begin{enumerate}[label=\roman*)]
    \item We prove that the concept of smooth sensitivity cannot be utilized along with the exponential mechanism. Therefore, we extend the smooth sensitivity, originally defined for numerical data, to the data selection setting;
    \item We propose the \metname (\met), a differentially private data selection algorithm that applies our extended notion of smooth sensitivity;
    \item We provide differential privacy guarantees for \metname, along with theoretically rigorous utility guarantees showing that \metname is never worse than its competitors under mild conditions;
    \item We conducted an empirical comparison\footnote{The source code and other artifacts have been made available at \\\url{https://github.com/iagocc/smooth-noisy-max}} of \metname with competing methods across three applications: percentile selection, greedy decision trees, and random forests. Our findings indicate that \met consistently outperforms state-of-the-art methods in terms of accuracy and expected error.
\end{enumerate}

This paper is structured as follows: Section 2 provides basic definitions regarding DP. Section 3 reviews the prior art on private selection. Section 4 presents the \metname algorithm. Section 5 applies \metname to percentile selection. Section 6 explores a private decision tree approach. Section 7 discusses a novel random forest algorithm using \metname. Finally, Section 8 concludes the paper with future directions. \looseness=-1

%% file: sections/2-background.tex
\section{Preliminaries}
\label{cap:bg}
Let database $\db{x}$ be a set of records drawn from a universe \( \univ \), and $f$ a query over $\db{x}$.
In differential privacy, the goal is to ensure that the outcome of a computation/algorithm, denoted by $\mathscr{A}$, does not reveal much sensitive information about any individual in a database. At the same time, the algorithm $\mathscr{A}$ ensures data processing without disclosing individual information, even if an adversary has almost complete knowledge of all other individuals in the database. Differential privacy uses a randomized algorithm, i.e., a mechanism that adds controlled noise to the data, and it is based on a privacy budget parameter, typically denoted as $\varepsilon$, representing the desired level of privacy protection. We formalize the database as a \emph{multiset} of records of $\DB$. Therefore, the distance between two databases can be determined by counting the records that differ between them. More specifically, this distance is quantified using the symmetric difference of two sets, denoted as $d(\db{x}, \db{y}) = |\db{x} \oplus \db{y}|$.

\begin{definition}[$(\varepsilon, \delta)$-Differential privacy \cite{dwork2014algorithmic}]
     A randomized algorithm $\mathscr{A}$ satisfies $(\varepsilon, \delta)$-differential privacy if, for any two databases $\db{x}$ and $\db{y}$ that differ in at most one record, and for any possible output $S \subseteq \Y$ over the outcome space $\Y$ of the algorithm
    $$
    Pr[\mathscr{A}(\db{x}) \in S] \leq e^\varepsilon Pr[\mathscr{A}(\db{y}) \in S] + \delta
    $$
    where $Pr[\cdot]$ stands for probability of an event. When $\delta = 0$, the algorithm is $\varepsilon$-differentially private. We refer to \(\varepsilon\)-differential privacy as pure differential privacy. Conversely, \((\varepsilon, \delta)\)-differential privacy, where \(\delta > 0\), is referred to as approximate differential privacy.
    \label{def:dp}
\end{definition}

An alternative interpretation of differential privacy is presented in Remark 3.1 of \citet{dwork2014algorithmic}, utilizing the concept of $\delta$-approximate max divergence. This perspective reformulates differential privacy in terms of distributional distance measures.

\begin{definition}[$\delta$-Approximate Max Divergence \cite{dwork2014algorithmic}] \label{def:dmaxdiv}
    The $\delta$-Approximate Max Divergence between two random variables $X$ and $Y$ taking values from the same domain is defined to be:
    \[ D_{\infty}^\delta (X||Y) = \max_{S \subseteq \Y \, : \, Pr[X \in S] \geq \delta} \left[  \log \left(  \frac{Pr[X \in S] - \delta}{Pr[Y \in S]} \right) \right] \]
\end{definition}

\begin{lemma}[Approx. Differential Privacy \cite{dwork2014algorithmic}]\label{def:adpremark}
    Note that a mechanism $\mathscr{A}$ is $(\varepsilon, \delta)$-differentially private if and only if on every two neighboring databases $\db{x},\db{y}: D_{\infty}^\delta (\mathscr{A}(\db{x})||\mathscr{A}(\db{y})) \leq \varepsilon$ and $D_{\infty}^\delta (\mathscr{A}(\db{y})||\mathscr{A}(\db{x})) \leq \varepsilon$.
\end{lemma}

The quantity of noise introduced is proportional to the global sensitivity of the query. Global sensitivity, denoted by $\Delta f$, quantifies the maximum change in a function's $f$ output when a single individual's data is modified, reflecting the largest difference between outputs for databases differing by one record. 

\begin{definition}[Global sensitivity \cite{dwork2014algorithmic}] \label{def:global_sens}
    The global sensitivity of a function $f: \DB \rightarrow \mathbb{R}$ is defined:
    \[
        \Delta f = \max_{\substack{\db{x},\db{y} \in \DB \\ d(\db{x},\db{y}) \leq 1}} \lvert f(\db{x}) - f(\db{y}) \rvert
    \]
\end{definition}

However, its practical utility is often limited due to excessive noise generation, as the Laplace mechanism's scale parameter is \(\frac{\Delta f}{\varepsilon}\) \cite{dwork2014algorithmic}, leading to high noise levels for functions like $k$-clique counting \cite{zhang2015private} and median queries \cite{nissim2007smooth}. Local sensitivity, denoted by $LS_f(\cdot)$, is a database-specific measure of the maximum change in output resulting from individual data modifications. Consequently, employing instance-specific sensitivity can help reduce the amount of noise introduced.

It is crucial to highlight that the global sensitivity is the maximum local sensitivity over all databases, $\Delta f = \max_{\db{x} \in \DB} LS_f(\db{x})$. Nonetheless, using the local sensitivity, instead of global, would reduce the amount of noise produced by the random algorithm so much that it would not satisfy the differential privacy definition \cite{nissim2007smooth}.

To address the problem of achieving differential privacy for numerical queries with instance-based sensitivity, the work of \citeauthor{nissim2007smooth} \cite{nissim2007smooth} proposed the smooth sensitivity framework, which smooths the local sensitivity at a distance $t$.

The local sensitivity at a distance $t$ measures the maximum local sensitivity $LS_f$ over all databases up to the distance $t$ from $\db{x}$, i.e., up to $t$ modifications on the database $\db{x}$. It is important to note that it is a generalization of the local sensitivity $LS_f(\db{x}, 0) = LS_f(\db{x})$, a particular case when the distance is set to $0$.

\begin{definition}[Local sensitivity at distance $t$ \cite{nissim2007smooth}]
    For a query $f: \DB \rightarrow \mathbb{R}^k$ and a database $\db{x} \in \DB$, the local sensitivity of $f$ at $\db{x}$ at distance $t$ is defined as:
    $$
        LS_f(\db{x}, t) = \max_{\substack{\db{y} \in \DB \;|\; d(\db{x},\db{y}) \leq t}} LS_f(\db{y})
    $$
\end{definition}

The sensitivity measure itself may inadvertently disclose individual information. Moreover, adjusting noise based on local sensitivity may risk potential data leakage. To determine the appropriate noise magnitude, the work by \citeauthor{nissim2007smooth} \cite{nissim2007smooth} utilizes a smooth upper bound on local sensitivity. Specifically, they define a function \( S \) that not only provides an upper limit on \( LS_f \) across all points but also ensures that \( \ln(S(\cdot)) \) maintains low sensitivity.

\begin{definition}[Smooth bound \cite{nissim2007smooth}] \label{def:sbound}
    For $\beta > 0$, a function $S: \DB \rightarrow \mathbb{R}^+$ is a $\beta$-smooth upper bound on the local sensitivity of a function $f$ if it satisfies the following requirements:
    $$
      \begin{aligned}
          \forall \db{x} \in \DB :\,                       & S(\db{x}) \geq LS_f(\db{x}) \\
          \forall \db{x},\db{y} \in \DB, d(\db{x},\db{y}) \leq 1 :\,     & S(\db{x}) \leq e^\beta S(\db{y})
      \end{aligned}
    $$
\end{definition}

\begin{definition}[Smooth sensitivity \cite{nissim2007smooth}] \label{sec:smooth}
    For $\beta > 0$, the $\beta$-smooth sensitivity of $f$ is:
    $$
        \SSENS_{f, \beta}(\db{x}) = \max_{t=0,1,\ldots, \lvert \db{x} \rvert} \left( e^{- t \beta } \cdot LS_f(\db{x}, t) \right)
    $$
\end{definition}

The smooth sensitivity $\SSENS_{f, \beta}$ is the smallest function to satisfy the smooth bound requirements (Definition \ref{def:sbound}) \cite{nissim2007smooth}. The smooth sensitivity adjusts the contribution of the local sensitivity based on the distance between a database and \(\db{x}\). The $\beta$ parameter, which serves as a smoothing factor, is strategically chosen to mitigate inadvertent data disclosure risks that may arise when employing local sensitivity directly. The global sensitivity $\Delta f$ is also a smooth upper bound on the local sensitivity, i.e., the global sensitivity satisfies the Definition \ref{def:sbound}. \looseness=-1

\begin{corollary}[Smooth sensitivity upper bound] \label{cor:upper_smooth}
    For a query $f$, a database $\db{x}$, the global sensitivity $\Delta f$ is an upper bound of smooth sensitivity $\SSENS_{f, \beta}$ i.e., $\SSENS_{f, \beta}(\db{x}) \leq \Delta f$.
\end{corollary}

Mechanisms that the addition of noise is proportional to the smooth sensitivity are contingent upon whether the noise distribution meets the criteria necessary for achieving differential privacy, i.e., $(\alpha, \beta)$-admissibility.

\begin{definition}[Admissible Noise Distribution \cite{nissim2007smooth}] \label{def:adm}
    A probability distribution on $\mathbb{R}^k$, given by a density function $h$, is $(\alpha, \beta)$-admissible if, for $\alpha = \alpha(\varepsilon, \delta)$, $\beta = \beta(\varepsilon, \delta)$, the following two conditions hold for all $\Delta \in \mathbb{R}^k$ and $\lambda \in \mathbb{R}$ satisfying $\|\Delta\|_1 \leq \alpha$ and $|\lambda| \leq \beta$, and for all measurable subsets $S \subseteq \mathbb{R}^k$.
    \begin{enumerate}[label=(\roman*)]
        \item \textit{(Sliding)} \quad\quad $\underset{Z \sim h}{Pr}[Z \in S] \leq e^{\frac{\varepsilon}{2}} \underset{Z \sim h}{Pr}[Z \in S + \Delta] + \frac{\delta}{2}$
        \item \textit{(Dilation)} \quad\; $\underset{Z \sim h}{Pr}[Z \in S] \leq e^{\frac{\varepsilon}{2}} \underset{Z \sim h}{Pr}[Z \in S\cdot e^\lambda] + \frac{\delta}{2}$
    \end{enumerate}
\end{definition}

%% file: sections/3-priv_selection.tex
\section{Private Selection}
This section covers works on private selection from the literature, as this paper specifically addresses the private selection problem. Private selection refers to selecting the best item, or outcome, option from a set of possible outputs while ensuring the individual's data privacy. Formally, we want to build a private algorithm for a query $f: \DB \rightarrow \OUT$ where all possible outcomes for $f$ are discrete, e.g., categorical values. In the private selection setting is necessary a utility function $u: \DB \times \OUT \rightarrow \mathbb{R}$ that maps a database $\db{x}$ and an output $r \in \OUT$ to a utility score $u(\db{x}, r)$. This utility function is application-based, and the higher the utility values are, the better the outcome is for the database.

We now review the prior art on differentially private data selection algorithms, which comprises the well-established exponential \cite{mcsherry2007mechanism} and the report-noisy-max  \cite{dwork2014algorithmic} mechanisms, as well as the recently proposed permute-and-flip \cite{mckenna2020permute} and the local dampening  \cite{farias2023local} mechanisms. \looseness=-1

\subsubsection*{Exponential Mechanism}
The exponential mechanism in the private selection setting is the \textit{de facto} standard. It samples possible outputs from $\OUT$ with a probability that grows exponentially with their utility function $u$.

\begin{definition}[Exponential Mechanism \cite{mcsherry2007mechanism}]
    The exponential mechanism $\MEXP_{\score, \varepsilon}(\db{x}, r)$ selects an outcome $r \in \OUT$ as follows:
    $
        \MEXP_{\score, \varepsilon}(\db{x}, r) \propto \exp\left({\frac{\varepsilon \score(\db{x}, r)}{2\Delta \score}} \right)
    $,
    where $\Delta \score$ is the global sensitivity of the utility function $\score$.
\end{definition}

\citeauthor{mcsherry2007mechanism} \cite{mcsherry2007mechanism} showed that the exponential mechanism satisfies $\varepsilon$-DP through global sensitivity, i.e., using noise with scale modulated by the global sensitivity. \looseness=-1

\begin{definition}[Global Sensitivity \cite{mcsherry2007mechanism}]
    Let $\score : \DB \times \OUT \rightarrow \mathbb{R}$ be a utility function that maps a pair of a database and an outcome to a score. The global sensitivity of $\score$ is:
    $$
        \Delta \score = \max_{r \in \OUT} \max_{\substack{\db{x}, \db{y} \in \DB \,|\, d(\db{x},\db{y}) \leq 1}} \lvert \score(\db{x}, r) - u(\db{y}, r) \rvert    
        $$
    
\end{definition}

\subsubsection*{Permute-and-flip}
Another private selection algorithm, called Permute-and-Flip, was proposed by \citeauthor{mckenna2020permute} \cite{mckenna2020permute}. 
The algorithm works by iterating over the set of outcomes $\OUT$ in a random order, and for each element $r$, it flips a biased coin with a certain probability. If the flipped coin lands tails, then $r$ is removed from all possible outcomes. Otherwise (if it lands heads), $r$ is the returned outcome for the mechanism. The likelihood of obtaining heads follows an exponential pattern concerning the quality score, thereby boosting the mechanism to produce results with superior quality scores. While the permute-and-flip algorithm achieves $\varepsilon$-differential privacy, this guarantee only applies under the global sensitivity $\Delta u$. For problems with high global sensitivity, the algorithm might suffer from reduced accuracy, as well as the exponential mechanism.

The work proves that the expected error of permute-and-flip is never worse than that of the exponential mechanism. Moreover, it shows that the exponential mechanism can be viewed as a rejection sampling algorithm that samples uniformly from the outcome set $\OUT$ with replacement. On the other hand, the permute-and-flip works like an exponential mechanism but sampling without replacement from $ \OUT $.

\subsubsection*{Report-noisy-max}
Nevertheless, another private selection algorithm is proposed by \citeauthor{dwork2014algorithmic} \cite{dwork2014algorithmic} called the report-noisy-max. 
The algorithm adds independent noise to each outcome utility score and returns the outcome with the highest noisy score. \citeauthor{dwork2014algorithmic} \cite{dwork2014algorithmic} proposes the algorithm with noise sampled by the Laplace distribution. However, the algorithm can be generalized to other noise distributions, such as the Gumbel and Exponential distributions. \looseness=-1
%

This algorithm is a broad private selection method. Specifically, the report-noisy-max with the Exponential distribution, denoted by $\RNMEXP$, samples noise from $\texttt{Expo}\left( {\varepsilon}/{2\Delta \score} \right)$, and this version also has strong utility guarantees shown by the Theorem \ref{the:rnm_util}. It is identical to permute-and-flip \cite{ding2021permute}. Moreover, the report-noisy-max with the Gumbel distribution $\texttt{Gumbel}\left({2\Delta \score}/{\varepsilon}\right)$ is identical to the exponential mechanism \cite{durfee2019practical}. Nevertheless, the report-noisy-max only holds the differential privacy requirements under the global sensitivity of the utility function, which might lead to poor accuracy under certain scenarios \cite{zhang2015private, gonem2018smooth, DBLP:journals/corr/abs-2003-00505, DBLP:conf/nips/BunS19}. \looseness=-1


The report-noisy-max inadvertently discards information \cite{ding2023free}. More precisely, without incurring any supplementary privacy costs, it can disclose an estimate of the difference between the two largest noisy utility values.

\begin{theorem} \label{the:rnm_util}
     Consider the report-noisy-max with exponential distribution $\RNMEXP$ algorithm. Let $\db{x} \in \DB$ be a fixed database, $\xi$ be the error and $\OUT$ be the set of all possible outcomes. Then, for a given $t > 0$, the following inequalities hold:
	\begin{enumerate}[label=(\roman*)]
		\item $Pr\left[\xi(\RNMEXP, \db{x}) \geq \frac{2 \Delta u \, (\ln(|\OUT|) + t)}{\varepsilon} \right] \leq e^{-t};$
		\item $\mathbb{E}\left(\xi(\RNMEXP, \db{x})\right) \leq \frac{2 \Delta u \, (\ln(|\OUT|) + 1)}{\varepsilon}.$
	\end{enumerate}
\end{theorem}


\subsubsection*{Local Dampening Mechanism}
In specific scenarios, the global sensitivity may not be suitable because the global sensitivity is large and the signal-to-sensitivity ratio (i.e. $\nicefrac{\text{utility}}{\text{sensitivity}}$) is too low, implying inaccurate results. To address this issue, the local dampening mechanism \cite{farias2023local} designs an instance-based sensitivity to work along with a novel mechanism based on the exponential one. It also proposes new adapted versions of the local sensitivity at a distance $t$ to the private selection setup.

\begin{definition}[Local Sensitivity for private selection \cite{farias2023local}]
    Let $u : \DB \times \OUT \rightarrow \mathbb{R}$ be a utility function that maps a pair of a database and an outcome to a score. The local sensitivity is defined as:
    $$
        LS_\score(\db{x}) = \max_{r \in \OUT} \; \max_{\substack{\db{y} \in \DB \\ d(\db{x},\db{y}) \leq 1}} \left| \score(\db{x}, r) - \score(\db{y}, r) \right|
    $$
\end{definition}

\begin{definition}[Local Sensitivity at distance $t$ for private selection \cite{farias2023local}] \label{def:lst}
    Let $u : \DB \times \OUT \rightarrow \mathbb{R}$ be a utility function that maps a pair of a database and an outcome to a score. The local sensitivity of a function $\score$ for the database $\db{x}$ at distance $t$ is defined as:
    $$
        LS_\score(\db{x}, t) = \max_{\substack{\db{y} \in \DB \\ d(\db{x},\db{y}) \leq t}} LS_\score(\db{y})
    $$
\end{definition}

Whereas the local sensitivity at distance $t$ provides an overview of the utility $\score$ variation in its neighborhood, it lacks in granting more information about the utility function $\score$ with a specific outcome $r$ in its neighborhood. Therefore, the paper \cite{farias2023local} proposes a novel generalization of local sensitivity called the element local sensitivity. It measures the sensitivity of a utility function $\score$ for a specific outcome $r$ at a distance $t$.


The computation of the element's local sensitivity is only sometimes feasible because it could be NP-hard. Therefore, the paper proposes a definition that represents an heuristic to compute an upper bound to the element's local sensitivity, named admissible function $\delta_\score$.

The local dampening attenuates the utility function in a specific way to make the signal-to-sensitivity ratio larger. This function is called $D_{\score, \delta^\score}$ and uses an admissible function $\delta^\score$ that provides a  dampened and scaled version of the original utility function. And finally, the local dampening mechanism $\MLD_{\score,\varepsilon, \delta_\score}$ selects an element $r \in \OUT$ with probability proportional to $\exp \left( \frac{\varepsilon \cdot D_{\score, \delta_\score}(\db{x},r)}{2} \right)$.


The local dampening mechanism satisfies $\varepsilon$-differential privacy if $\delta$ is admissible. It also performs at least equal to the exponential mechanism when the sensitivity function meets specific scenarios, such as stability. However, there are a few caveats to the local dampening mechanism, particularly related to the inversion problem, the necessity for stability, and the time complexity.

%% file: sections/4-method.tex
\section{\metname}
\label{sec:met}

This paper introduces \metname (\met), an algorithm that tackles the differentially private selection problem. The proposed algorithm is inspired by the report-noisy-max. \met offers significant advantages over the existing methods, such as simplicity, ease of implementation, and accuracy performance. In particular, our novel approach adopts an instance-based sensitivity instead of global sensitivity, as the latter can often be excessively large, leading to a low signal-to-sensitivity ratio (i.e., \(\nicefrac{\text{utility}}{\text{sensitivity}}\)) and, consequently, inaccurate results.

More precisely, \met applies the smooth sensitivity.

%
\begin{definition}[Smooth sensitivity, adapted from \citeauthor{nissim2007smooth} \cite{nissim2007smooth}]
	For $\beta > 0$, the $\beta$-smooth sensitivity of the utility function $u$ is:
    \[
        \SSENS_{u, \beta}(\db{x}) = \max_{t=0,1,\ldots, \lvert \db{x} \rvert} \left( e^{- t \beta } \cdot LS_u(\db{x}, t) \right)
    \]
    \label{def:mysmooth}
\end{definition}%
The smooth sensitivity attenuates the local sensitivity (Definition \ref{def:lst}) based on the distance from $\db{x}$. Applying an instance-based sensitivity, such as smooth sensitivity, within a private selection algorithm is not always feasible for differential privacy. For instance, the exponential mechanism can not be used directly with the smooth sensitivity (see Theorem \ref{the:sam}).
On the other hand, the proposed \metname algorithm can take advantage of the smooth framework and consequently decrease the signal-to-sensitivity ratio of the method. Additionally, it can keep the same differentially private guarantees of the standard report-noisy-max and ensures better accuracy.

\met adds noise proportional to a smooth upper bound on the local sensitivity (e.g. smooth sensitivity $\SSen$) to its utility value for each possible outcome $r$ for the query $f$ at database $\db{x}$, i.e., $\score(\db{x},r)$.
The noise, expressed by a random variable $Z$, is drawn from an $(\alpha, \beta)$-admissible probability density function (Definition \ref{def:adm}). For the sake of simplicity, we refer to $\SSen$ as $\SSENS$. This procedure is explained in Algorithm \ref{alg:met}.
%
%
\begin{algorithm}[!htbp]
  \caption{\metname Algorithm}\label{alg:met}
  \For{$r \in \OUT$} {
    $\tilde{u}(\db{x}, r) \gets u(\db{x}, r) + \frac{2\SSENS(\db{x})}{\alpha} \cdot Z$\;
  }
  \Return $\argmax_{r \in \mathcal{R}} \tilde{u}(\db{x}, r)$
\end{algorithm}
\subsection{Privacy Guarantees}
In Theorem \ref{theo:met-dp}, we prove that the \metname algorithm ensures $(\varepsilon, \delta)$-differential privacy.

\begin{theoremEnd}[end, restate,category=priv,text link={\noindent \textit{Proof.} See proof on appendix \ref{apx:priv}}]{theorem}
    The \metname $\LEXP$ algorithm is $(\varepsilon, \delta)$-differentially private if $h$ is an $(\alpha, \beta)$-admissible noise probability density function, and $Z$ a random variable sampled according to $h$.
    \label{theo:met-dp}
\end{theoremEnd}
\begin{proofE}
        Consider two neighbor databases $\db{x}$ and $\db{y}$.
        Fix any $i \in \OUT$ and let $\vec{z}_{i} = \{ z_1, \ldots, z_{|\OUT|}\} \backslash \{z_i\}$ be the fixed noises for all outputs except the \textit{i}th output.
        We will argue for each $\vec{z}_{i}$ independently, similarly to what was done by \citet{dwork2014algorithmic} (Claim 3.9).
        For simplicity of notation, denote $N(\db{x}) = \nicefrac{2\SSen(\db{x})}{\alpha}$, and the \metname as $\RLEXP$.
        Then, the probability of $i \in \OUT$ being the output of the algorithm, given the noises $\vec{z}_{i}$, is
        \[
            Pr[\RLEXP(\db{x}) = i | \vec{z}_i] = Pr\left[u(\db{x}, i) + N(\db{x})\cdot Z
                \geq \max_{j \in \OUT;j\neq i } \left\{ u(\db{x}, j) + z_j \right\} \right].
        \]
        Let $\tilde{u}_* = \max\limits_{j \in \OUT;j\neq i}\{ u(\db{x}, j) + z_j \}$ and $\tilde{u}'_* = \max\limits_{j \in \OUT;j\neq i}\{ u(\db{y}, j) + z_j \}$. Then:
        \[
            Pr[\RLEXP(\db{x}) = i | \vec{z}_i] = Pr\left[Z \geq \frac{\tilde{u}_* - u(\db{x}, i)}{N(\db{x})} \right].
        \]
        For the sake of simplicity, define $g(i) = \frac{\tilde{u}_* - u(\db{x}, i)}{N(\db{x})}$ and $g'(i) = \frac{\tilde{u}'_* - u(\db{y}, i)}{N(\db{y})}$, so that:
        \[
            Pr[\RLEXP(\db{x}) = i | \vec{z}_i] = Pr\left[Z \geq g(i) \right].
        \]
        Using the definition \ref{def:dmaxdiv} for neighboring databases $\db{x}, \db{y}$, and letting $Z_X \sim \RLEXP(\db{x})$, $Z_Y \sim \RLEXP(\db{y})$:
        \[
            D_{\infty}^\delta (Z_X || Z_Y) = \max_{\substack{S \subseteq \OUT \, : \\ Pr[Z_X \in S] \geq \delta}} \left[  \log \left(  \frac{Pr[Z_X \in S] - \delta}{Pr[Z_Y \in S]} \right) \right].
        \]
        As our algorithms draws results from the discrete set of outputs, we can write:
        \begin{align*}
            D_{\infty}^\delta (Z_X || Z_Y) &= \max_{\substack{S \subseteq \OUT \, : \\ Pr[Z_X \in S] \geq \delta}} \left[  \log \left( \frac{\sum\limits_{r \in S} Pr[Z_X = r] - \delta}{\sum\limits_{r \in S} Pr[Z_Y = r]} \right) \right], \\
            &= \max_{\substack{S \subseteq \OUT \, : \\ Pr[Z_X \in S] \geq \delta}} \left[  \log \left(  \frac{\sum\limits_{r \in S} \int Pr[Z_X = r | \vec{z_r}]Pr[\vec{z_r}]  d\vec{z_r} - \delta}{\sum\limits_{r \in S} \int Pr[Z_Y = r | \vec{z_r}] Pr[\vec{z_r}] d\vec{z_r}} \right) \right],\\
            &= \max_{\substack{S \subseteq \OUT \, : \\ Pr[Z_X \in S] \geq \delta}} \left[  \log \left(  \frac{\sum\limits_{r \in S} \int Pr\left[Z_X \geq g(r) \right] Pr[\vec{z_r}] d\vec{z_r} - \delta}{\sum\limits_{r \in S} \int Pr\left[Z_Y \geq g'(r) \right] Pr[\vec{z_r}] d\vec{z_r}} \right) \right].
        \end{align*}
        Since $Z \sim h$ and $h$ is admissible, we can use the sliding property:
        \begin{align*}
            D_{\infty}^\delta (Z_X || Z_Y) 
            &\leq \max_{\substack{S \subseteq \OUT \, : \\ Pr[Z_X \in S] \geq \delta}} \left[  \log \left(  \frac{\sum\limits_{r \in S} \int Pr\left[Z_X \geq g(r) - g(r) + \frac{\tilde{u}'_* - u(\db{y}, r)}{N(\db{x})} \right] \cdot e^{\frac{\e}{2}} Pr[\vec{z_r}] d\vec{z_r} + \frac{\delta}{2} - \delta}{\sum\limits_{r \in S} \int Pr\left[Z_Y \geq g'(r) \right] Pr[\vec{z_r}] d\vec{z_r}} \right) \right], \\
            &\leq \max_{\substack{S \subseteq \OUT \, : \\ Pr[Z_X \in S] \geq \delta}} \left[  \log \left(  \frac{\sum\limits_{r \in S} \int Pr\left[Z_X \geq \frac{\tilde{u}'_* - u(\db{y}, r)}{N(\db{x})} \right] \cdot e^{\frac{\e}{2}} Pr[\vec{z_r}] d\vec{z_r} -\frac{\delta}{2}}{\sum\limits_{r \in S} \int Pr\left[Z_Y \geq g'(r) \right] Pr[\vec{z_r}] d\vec{z_r}} \right) \right].
        \end{align*}
        The first inequality results from the sliding property since $h$ is admissible. Notice that this property can be applied above because, by the properties of smooth and local sensitivities, and since $\db{x}$ and $\db{y}$ are neighbors:
        \begin{align*}
            - g(r) + \frac{\tilde{u}'_* - u(\db{y},r)}{N(\db{x})} = \frac{- \tilde{u}_* + u(\db{x}, r) + \tilde{u}'_* -u(\db{y}, r)}{N(\db{x})} &= \alpha \frac{-u(\db{y}, r) + u(\db{x}, r) - \tilde{u}_* + \tilde{u}'_*}{2\SSen(\db{x})},\\
            &\leq \frac{\alpha}{2LS(\db{x})} \left(\underbrace{u(\db{x}, r) - u(\db{y}, r)}_{\leq LS(\db{x})} + \underbrace{\tilde{u}'_* - \tilde{u}_*}_{\leq LS(\db{x})}\right),\\
            &\leq \alpha \frac{2LS(\db{x})}{2LS(\db{x})} = \alpha.
        \end{align*}
        Further we can apply the dilation property since $h$ is admissible and $\ln \frac{N(\db{x})}{N(\db{y})} = \ln \frac{\SSen(\db{x})}{\SSen(\db{y})} \leq \beta$ (see Definition \ref{def:sbound}):
        \begin{align*}
        D_{\infty}^\delta (Z_X || Z_Y) 
        &\leq \max_{\substack{S \subseteq \OUT \, : \\ Pr[Z_X \in S] \geq \delta}} \left[  \log \left(  \frac{\sum\limits_{r \in S} \int Pr\left[Z_X \geq \frac{\tilde{u}'_* - u(\db{y}, r)}{N(\db{x})} \cdot \frac{N(\db{x})}{N(\db{y})} \right] \cdot e^{\frac{\e}{2}}\cdot e^{\frac{\e}{2}} Pr[\vec{z_r}] d\vec{z_r} + \frac{\delta}{2} - \frac{\delta}{2}}{\sum\limits_{r \in S} \int Pr\left[Z_Y \geq g'(r) \right] Pr[\vec{z_r}] d\vec{z_r}} \right) \right], \\
        &= \max_{\substack{S \subseteq \OUT \, : \\ Pr[Z_X \in S] \geq \delta}} \left[  \log \left(  \frac{\sum\limits_{r \in S} \int Pr\left[Z_X \geq \frac{\tilde{u}'_* - u(\db{y}, r)}{N(\db{y})} \right] \cdot e^{\e} Pr[\vec{z_r}] d\vec{z_r} }{\sum\limits_{r \in S} \int Pr\left[Z_Y \geq g'(r) \right] Pr[\vec{z_r}] d\vec{z_r}} \right) \right], \\
        &= \max_{\substack{S \subseteq \OUT \, : \\ Pr[Z_X \in S] \geq \delta}} \left[  \log \left(  \frac{e^{\e} \cdot \sum\limits_{r \in S} \int Pr\left[Z_X \geq g'(r) \right] Pr[\vec{z_r}] d\vec{z_r}}{\sum\limits_{r \in S} \int Pr\left[Z_Y \geq g'(r) \right] Pr[\vec{z_r}] d\vec{z_r}} \right) \right], \\
        &= \e.
        \end{align*}
        By symmetry, we can also prove that $D_{\infty}^{\delta}(Z_Y \| Z_X) \leq \varepsilon$. Then, by Definition \ref{def:adpremark}, we conclude that \met is $(\varepsilon, \delta)$-differentially private.
\end{proofE}

\begin{corollary}
    The \metname \( \LEXP \) algorithm with sampled noise from the Student's T distribution is \(\varepsilon\)-differentially private.
    By scaling the Student's T distribution in accordance with the smooth sensitivity, pure differential privacy is assured \cite{avg2019}.
\end{corollary}

\begin{corollary}
    The \metname \( \LEXP \) algorithm with sampled noise from the Laplace distribution is \((\varepsilon, \delta)\)-differentially private, when $\beta$ parameter is defined by $\nicefrac{\varepsilon}{2\log\left( \nicefrac{2}{\delta} \right)}$ \cite{nissim2007smooth}.
\end{corollary}

\begin{corollary}
    The \metname \( \LEXP \) algorithm with sampled noise from the Laplace Log-Normal (\texttt{LLN}($\sigma$)) distribution is \((\varepsilon, \delta)\)-differentially private \cite{avg2019}, when $\alpha$ parameter is defined as $e^{-\nicefrac{3}{2}\sigma^2}\left(\varepsilon - \nicefrac{\beta}{\sigma}\right)$.
\end{corollary}

We can also improve the noise addition under the monotonicity property. The utility function $u$ is monotonic in the database if adding an element to the database cannot cause the value of the function to decrease, e.g., counting queries.

\begin{corollary}\label{col:met_mono}
  When the utility function $u$ is monotonic in the database, then the \metname $\LEXP$ scales the noise only by a factor of $\frac{\SSENS(\db{x})}{\alpha}$.
\end{corollary}

\subsection{Utility Analysis}

A significant characteristic of the \metname algorithm is that it provides strong utility guarantees. Given a database $\db{x}$, we can find the error bound of the private algorithm by a specific parameter $t$. The algorithm's accuracy is assessed based on the largest utility score $u^* = \max_{r \in \OUT} u(\db{x},r)$. It will be highly unlikely that the returned element $r$ has a utility score significantly less than $O\left( u^* - (\nicefrac{\SSen(\db{x})}{\varepsilon}) \ln|\OUT| \right)$ when the noise distribution is Laplace.
\begin{theoremEnd}[end, restate,category=util,text link={\noindent \textit{Proof.} See proof on appendix \ref{apx:util}}]{lemma}
    \label{lemma:tb_exp}
	Given a fixed database $\db{x} \in \DB$, for the \metname $\RLEXP$ algorithm with a standard Laplace distribution as noise function and any $t>0$, the error $\xi(\RLEXP, \db{x})$ satisfies
	\begin{align*}
		Pr[\xi(\RLEXP, \db{x}) \geq t] \leq |\OUT| \exp\left( - \frac{\varepsilon t}{4 \SSen(\db{x})} \right).
	\end{align*}
\end{theoremEnd}

\begin{proofE}
    Define $\score^*(\db{x}) = \max_{r \in \OUT} \score(\db{x}, r)$, so for each possible outcome $r \in \OUT$, the error can be written as $\xi(\RLEXP, \db{x}) = u^*(\db{x}) - u(\db{x}, r)$. Thus, for $t > 0$:
    \begin{equation}\label{eq:lema-4.7}
        Pr[\xi(\RLEXP, \db{x}) \geq t] = Pr[\score(\db{x}, \RLEXP(\db{x})) \leq \score^*(\db{x}) - t].
    \end{equation}
    For simplicity of notation, define the following subsets of $\OUT$:
    \begin{enumerate}[label=(\roman*)]
		\item $\OUT_t = \{ r \in \OUT : \score(\db{x}, r) \leq \score^*(\db{x}) - t \}$;
		\item $\OUT_* = \{ r \in \OUT : \score(\db{x}, r) = \score^*(\db{x}) \}$.
	\end{enumerate}
    Also, consider the noisy utility $\nscore(\db{x}, r) = \score(\db{x}, r) + \left(\nicefrac{2\SSen(\db{x})}{\alpha} \right) \cdot z_r$, where $ z_r \sim \texttt{Lap}\left(0, 1 \right)$, and its maximal value $\nscore^* (\db{x}) = \max_{r \in \OUT} \nscore(\db{x}, r)$.
    Notice that the probability of the output being in $\OUT_t$ is the same probability of existing some element in $\OUT_t$ with the greatest noisy utility. This way, the probability expressed in Equation \ref{eq:lema-4.7} is equivalent to the probability of existing some $r \in \OUT_t$ such that $\nscore(\db{x}, r) = \nscore^*(\db{x})$. In other words, $\exists r \in \OUT_t : \nscore(\db{x}, r) = \nscore^*(\db{x})$.
    Then:
    \begin{alignat*}{3}
        Pr[\xi(\RLEXP, \db{x}) \geq t] &= Pr[\exists r \in \OUT_t : \nscore(\db{x}, r) = \nscore^*(\db{x})] 
        &= Pr[\cup_{r \in \OUT_t} [\nscore(\db{x}, r) = \nscore^*(\db{x})]] 
        &\leq \sum_{r\in \OUT_t}Pr[\nscore(\db{x}, r) = \nscore^*(\db{x})].
    \end{alignat*}
    Let $r'$ be the most probable output in $\OUT_t$. In this case, we can write:
    \begin{align*}
        Pr[\xi(\RLEXP, \db{x}) \geq t] &\leq |\OUT_t| \ Pr[\nscore(\db{x}, r') = \nscore^*(\db{x})]
        = |\OUT_t| \ Pr[\nscore(\db{x}, r') \geq \nscore^*(\db{x})]
        = |\OUT_t| \ Pr[\left(\nicefrac{2\SSen(\db{x})}{\alpha} \right) \cdot z_{r'} \geq \nscore^*(\db{x}) - \score(\db{x}, r')],\\
        & \leq \frac{|\OUT_t| \ Pr[z_{r'} \geq (\nscore^*(\db{x}) - \score(\db{x}, r')) \cdot \left(\nicefrac{\alpha}{2\SSen(\db{x})} \right) ]}{Pr[\RLEXP(\db{x}) \in \OUT_*]}.
    \end{align*}
    Notice that $Pr[\nscore(\db{x}, r') = \nscore^*(\db{x})] = Pr[\nscore(\db{x}, r') \geq \nscore^*(\db{x})]$, since $\nscore^*(\db{x})$ is the maximal noisy utility.
    Now, consider $r^* = \argmax_{r \in \OUT} \score(\db{x}, r)$ and $z_*$ as the noise associated with $r^*$. Thus, we can write:
    \begin{align*}
        Pr[\RLEXP(\db{x}) \in \OUT_*] &= Pr[\cup_{r \in \OUT_*} [\RLEXP(\db{x}) = r]],\\
        &= \sum_{r \in \OUT_*} Pr[\RLEXP(\db{x}) = r],\\
        &= |\OUT_*| \ Pr[\RLEXP(\db{x}) = r^*],\\
        &= |\OUT_*| \ Pr[z_* \geq (\nscore^*(\db{x}) - \score(\db{x}, r^*)) \cdot \left(\nicefrac{\alpha}{2\SSen(\db{x})} \right)].
    \end{align*}
    The equality above is valid because $u(\db{x}, r) = u^*(\db{x}) \ \forall r \in \OUT_*$, so the chance of any of them being the output depends only on the noise, resulting in independent events with equal probability. As a consequence:
    \begin{align*}
        Pr[\xi(\RLEXP, \db{x}) \geq t] \leq \frac{|\OUT_t| \ Pr[z_{r'} \geq (\nscore^*(\db{x}) - \score(\db{x}, r')) \cdot \left(\nicefrac{\alpha}{2\SSen(\db{x})} \right) ]}{|\OUT_*| \ Pr[z_* \geq (\nscore^*(\db{x}) - \score(\db{x}, r^*)) \cdot \left(\nicefrac{\alpha}{2\SSen(\db{x})} \right)]}.
    \end{align*}
    However, as $z_{r'}, z_* \sim \texttt{Lap}(0, 1)$, $\score(\db{x}, r^*) = \score^*(\db{x})$ and $\score(\db{x}, r') \leq \score^*(\db{x}) - t$:
    \begin{align*}
        \frac{|\OUT_t| \, Pr[z_{r'} \geq (\nscore^*(\db{x}) - \score(\db{x}, r')) \cdot \left(\nicefrac{\alpha}{2\SSen(\db{x})} \right)]}{|\OUT_*| \, Pr[z_* \geq (\nscore^*(\db{x}) - \score(\db{x}, r_*)) \cdot \left(\nicefrac{\alpha}{2\SSen(\db{x})} \right)]}
        &\ = \frac{\frac{|\OUT_t|}{2} \exp \left( -\frac{\alpha (\nscore^*(\db{x}) - \score(\db{x}, r'))}{2 \SSen(\db{x})} \right)}{\frac{|\OUT_*|}{2} \, \exp \left( -\frac{\alpha (\nscore^*(\db{x}) - \score(\db{x}, r_*))}{2 \SSen(\db{x})} \right)},\\
        & \ \leq \frac{|\OUT_t|}{|\OUT_*|} \, \frac{\exp \left( -\frac{\alpha (\nscore^*(\db{x}) - \score^*(\db{x}) + t)}{2 \SSen(\db{x})} \right)}{\exp \left( -\frac{\alpha (\nscore^*(\db{x}) - \score^*(\db{x}))}{2 \SSen(\db{x})} \right)},\\
        & \, = \frac{|\OUT|}{|\OUT_*|} \exp \left( -\frac{\alpha t}{2 \SSen(\db{x})} \right).
    \end{align*}
    We know that $\alpha = \frac{\varepsilon}{2}$ (see \citeauthor{nissim2007smooth}, Lemma 2.9 \cite{nissim2007smooth}). Then, from the result above, we can finally conclude that:
    $$
    Pr[\xi(\RLEXP, \db{x}) \geq t] \leq {|\OUT|} \exp \left( -\frac{\varepsilon t}{4 \SSen(\db{x})} \right).
    $$
\end{proofE}

\begin{theorem} \label{th:util_met}
    Let $\db{x} \in \DB$ be a fixed database. Then, for a given $t > 0$, the \metname $\RLEXP$ algorithm with {standard} Laplace noise distribution will have the following properties:
	\begin{enumerate}[label=(\roman*)]
		\item $Pr\left[\xi(\RLEXP, \db{x}) \geq \frac{4 \SSen(\db{x}) \, (\ln(|\OUT|) + t)}{\varepsilon} \right] \leq e^{-t}$;
		\item $\mathbb{E}\left(\xi(\RLEXP, \db{x})\right) \leq \frac{4 \SSen(\db{x}) \, (\ln(|\OUT|) + 1)}{\varepsilon}$.
	\end{enumerate}
\end{theorem}




The utility bounds presented by Theorem \ref{th:util_met} provide tools to compare and show that the \metname outperforms our related work, i.e., report-noisy-max, exponential mechanism, and permute-and-flip. Firstly, we analyze the utility of the \metname in contrast to the report-noisy-max with exponential noise, shown by Theorem \ref{th:met_gt_rlm}.

\begin{definition}
    An algorithm $\mathscr{A}$ is said to be \emph{never worse} than some other algorithm $\mathscr{B}$ when, given a dataset $\db{x}$:
    \begin{enumerate}[label=(\roman*)]
		\item $Pr\left[ \xi\left(\mathscr{A}, \db{x} \right) \geq t \right] \leq Pr\left[ \xi\left(\mathscr{B}, \db{x} \right) \geq t \right]$ for all $t \geq 0$;
		\item $\mathbb{E}[\xi\left(\mathscr{A}, \db{x} \right)] \leq \mathbb{E}[\xi\left(\mathscr{B}, \db{x} \right)]$.
	\end{enumerate}
\end{definition}

\begin{theorem} \label{th:met_gt_rlm}
    The \metname $\RLEXP$ with Laplace noise distribution is \emph{never worse} than $\RNMEXP$ report-noisy-max algorithm with exponential noise when $\SSen(\db{x}) \leq \frac{\Delta u}{2}$.
	\begin{proof}
		Using the lemma \ref{lemma:tb_exp}, we can obtain
		\begin{align*}
			Pr\left[ \xi\left(\RLEXP, x \right) \geq t \right] & \leq \frac{|\OUT|}{|\OUT_*|} \exp \left( -\frac{\varepsilon t}{4 \SSen(\db{x})} \right),
		\end{align*}
        where $|\OUT_*|$ is the set of outcomes with the highest utility value.
        We observe that when \(\SSen(\db{x}) \leq \frac{\Delta u}{2}\), then
		\begin{align*}
			\frac{|\OUT|}{|\OUT_*|} \exp \left( -\frac{\varepsilon t}{4 \SSen(\db{x})} \right) & \leq \frac{|\OUT|}{|\OUT_*|} \exp \left( -\frac{\varepsilon t}{2 \Delta u} \right), \\
			Pr\left[ \xi\left(\RLEXP, \db{x} \right) \geq t \right]                                 & \leq Pr\left[ \xi\left(\RNMEXP, \db{x} \right) \geq t \right].
		\end{align*}
		Furthermore, by the Theorem \ref{the:rnm_util}, the first statement (\textit{i}) holds.
		We want to prove the second statement (\textit{ii}). The expected error can be expressed in terms of complementary cumulative distribution function:
		\begin{align*}
			\mathbb{E}(\xi(\RLEXP,x)) = \int_0^{\infty} Pr[\xi(\RLEXP,x) \geq t]dt.
		\end{align*}
		We shown that $Pr\left[ \xi\left(\RLEXP, x \right) \geq t \right] \leq Pr\left[ \xi\left(\RNMEXP, x \right) \geq t \right]$, thus:
		\begin{align*}
			\begin{split}
				\mathbb{E}(\xi(\RLEXP,x)) - \mathbb{E}(\xi(\RNMEXP,x)) =\\
				\int_0^{\infty} Pr\left[ \xi\left(\RLEXP, x \right) \geq t \right] - Pr\left[ \xi\left(\RNMEXP, x \right) \geq t \right] dt \leq 0.
			\end{split}
		\end{align*}
		Thus, the \metname with Laplace noise distribution is never worse than the report-noisy-max algorithm with exponential noise when $\SSen(\db{x}) \leq \frac{\Delta u}{2}$.
	\end{proof}
\end{theorem}

\citeauthor{ding2021permute} \cite{ding2021permute} shows that the report-noisy-max with exponential noise is identical to the permute-and-flip, so as we know, by Theorem \ref{th:met_gt_rlm} the \metname is never worse than report-noisy-max algorithm with exponential noise when $\SSen(\db{x}) \leq \frac{\Delta u}{2}$, and consequently never worse than permute-and-flip mechanism under the same constraint.


The utility of our proposed method also outperforms the exponential mechanism and report-noisy-max with Gumbel noise when $\SSen(\db{x}) \leq \frac{\Delta u}{2}$. Since, by transitivity, \met surpass the permute-and-flip that exceeds the exponential mechanism \cite{mckenna2020permute}. Additionally, the exponential mechanism is identical to report-noisy-max with Gumbel noise \cite{durfee2019practical}, therefore the \metname is \emph{never worse} than report-noisy-max with Gumbel noise. All these results are expressed by Corollary \ref{col:met_all}.

\begin{corollary} \label{col:met_all}
    When $\SSen(\db{x}) \leq \frac{\Delta u}{2}$, \met $\RLEXP$ algorithm is \emph{never worse} than $\MPF$ permute-and-flip, $\MEXP$ exponential mechanism, and $\RNMGUM$ report-noisy-max with Gumbel noise.
\end{corollary}



The lack of utility bounds for the Local Dampening mechanism \cite{farias2023local} hampers a comparative assessment with our \metname. Nevertheless, the paper conducts an exhaustive empirical analysis in the subsequent sections.

Other commonly used admissible distributions include the Student's T and Laplace Log-Normal distributions \cite{avg2019}. The upper bounds of these utility functions may not readily suggest the better admissible noise distribution for a given problem. To aid in identifying a suitable distribution, one can use Chebyshev's inequality to compare the distributions by focusing on their variances.
For example, consider a comparison between the Laplace distribution and the Student's T distribution. The variance of the Laplace distribution is \(2b^2\), where \(b\) is the scale parameter. For the Student's T distribution with degrees of freedom \(d\), the variance is defined as \(\frac{d}{d-2}\) for \(d > 2\). According to Chebyshev's inequality, the Student's T distribution has a lower upper bound than the Laplace distribution when \(\frac{d}{d-2} < 2b^2\).
Similarly, we can compare the Laplace distribution with the Laplace Log-Normal distribution, which has a variance of \(2e^{2\sigma^2}\). When the scale parameter \(b\) satisfies \(b > e^{\sigma^2}\), the Laplace Log-Normal distribution presents a lower upper bound than the Laplace distribution according to Chebyshev's inequality.

%% file: sections/5-app_perc.tex
\section{Application --- Percentile Selection}
In this section, we address the percentile selection problem. The task is to return the \textit{p-th} percentile value from a set of real numbers.
\citeauthor{nissim2007smooth} \cite{nissim2007smooth} and \citeauthor{mckenna2020permute} \cite{mckenna2020permute} have dealt with similar tasks. \citeauthor{nissim2007smooth}'s work \cite{nissim2007smooth} addressed the challenge of privately releasing the numerical median of a dataset. \citeauthor{mckenna2020permute} \cite{mckenna2020permute} work attacks a similar problem, also for the median of the data, returning the bin value where it belongs.

\subsection{Problem Statement}

Given a dataset $\db{x}$ represented as a vector $[ x_1, \ldots, x_n ]$. For simplicity's sake, assume that every database $\db{x}$ is ordered such that $x_1 \leq \ldots \leq x_n$. Suppose that all the values lies in $[0, \Lambda]$, $0 \leq x_1 \leq \ldots \leq x_n \leq \Lambda$. The task is to return the percentile value where its element $x_i$ is as close as possible to the $p$-th percentile element.

\subsection{Private Mechanism and Sensitivity Analysis}

Following the task statement, various private selection algorithms are applicable, including the exponential mechanism, permute-and-flip, local dampening, and our proposed \metname variants. The algorithms select any value from a discrete subset of $[0, \Lambda]$, i.e., $\OUT \subseteq [0, \Lambda]$. We designed a utility function $u_p$ that assigns a maximum score of 1 when element $i$ matches the $p$-th element's value and a minimum score of 0 in all other cases; see Definition \ref{def:percutil}. \looseness=-1

\begin{definition}[Utility function for percentile selection problem]
    Consider a database $\db{x} \in \DB$, $n = |\db{x}|$, and $i \in \mathbb{Z}^{0+}$ a non-negative integer. The utility is defined as follows:
    \begin{equation*}
            u_p(\db{x}, i) = \begin{cases}
                1, & \quad \text{if } x_i = x_k, \text{where } k = {\left\lfloor \frac{p \cdot n}{100} \right\rfloor }; \\
                0, & \quad \text{otherwise.}
            \end{cases}
    \end{equation*}
    \label{def:percutil}
\end{definition}

Recall that the exponential mechanism and the permute-and-flip require the global sensitivity $\Delta {u_p}$, the local dampening requires the element local sensitivity, and the \metname expects the smooth sensitivity $\SSENS_{u_p}$.

\subsubsection*{Global Sensitivity}
The following example can show a worst-case scenario. For instance, let $p = 50$ implying that $k = \left\lfloor \frac{n}{2} \right\rfloor$. Let $\db{x}$ be a dataset with $n > 2$ and even, where $x_{<k} = 0$ and $x_{\geq k} = \Lambda$. Let $\db{y}$ be a neighboring dataset of $\db{x}$, where one element $x_{\geq k}$ has been removed. Thus we have $u(\db{x}, k) = 1$, and $u(\db{y}, k) = 0$ which implies that $u(\db{x}, k) - u(\db{y}, k) = 1$. Thus, $|u(\db{x}, r) - u(\db{y}, r)| \leq 1$ for all $r \in \OUT$, and any two neighboring datasets $\db{x}, \db{y}$.

\begin{proposition} Let $u_p$ be the utility function given by Definition \ref{def:percutil}. Then, the global sensitivity for percentile selection problem is 
$\Delta {u_p} = 1$. \looseness=-1
\end{proposition}

\subsubsection*{Local Sensitivity}
One must first compute the local sensitivity at a distance $t$ to compute the smooth sensitivity. Let $\db{x} \in \DB$ be a dataset, and $j = \min\left(\sum_{i=0}^{k-1} u(\db{x}, x_i), \sum_{i=k+1}^{n} u(\db{x}, x_i)\right)$ the smallest sequence of $p$-th value repetition length at left or right of position $k$. Thus, the $p$-th percentile value will remain the same until $2j + 1$ insertions and deletions from $\db{x}$ because of the floor function in the $k$ definition (Definition \ref{def:percutil}). Figure \ref{fig:ex-perc} provides an illustrative example of how $j$ is computed.

\begin{proposition}
Let $\db{x} \in \DB$ be a dataset, $j = \min\left(\sum_{i=0}^{k-1} u(\db{x}, x_i), \sum_{i=k+1}^{n} u(\db{x}, x_i)\right)$ as described above, and $u_p$ as in Definition \ref{def:percutil}. Then, the local sensitivity at distance $t$ for the problem of percentile selection is given by:
\begin{equation*}
        LS_{u_p}(\db{x}, t) = \begin{cases}
        1, & \quad \text{if } t \geq 2j + 1;\\
        0, & \quad \text{otherwise}.
    \end{cases}
\end{equation*}
\end{proposition}

\begin{figure}
    \centering
    \begin{align*}
        \db{x} &= \begin{bmatrix} 2\; 3\; 5\; \ldots\; \smash{\overbrace{5}^{k}}\; \ldots\; 5\; 6\; 7 \end{bmatrix} \\
        \boldsymbol{u}_p^{\db{x}}&= \begin{bmatrix} \smash{\underbrace{0\; 0\; 1\; \ldots\;}_{\boldsymbol{j}_{l}}} \textover[c]{1}{$\smash{\overbrace{5}^{k}}\;$} \; \smash{\underbrace{\ldots\; 1\; 0\; 0}_{\boldsymbol{j}_{g}}} \end{bmatrix}\\ \\
        j &= \max{\left( \textstyle\sum \boldsymbol{j}_{l}, \textstyle\sum \boldsymbol{j}_{g}\right)}
    \end{align*}
    \caption{The utility function \( u_p \) maps elements of \(\db{x}\) to a utility value. In this example, the \(x_k = 5\), and \(\boldsymbol{u}_p^{\db{x}}\) is the utility vector for the dataset \(\db{x}\). The subsets \( \boldsymbol{j}_l \) and \( \boldsymbol{j}_g \) partition the dataset into elements less than and greater than index \( k \), respectively. The final equation computes \( j \) as the maximum of the summed utility values, i.e., the number of elements that have the same value of \(x_k\) in each partition.}
    \label{fig:ex-perc}
\end{figure}

Now, it is possible to calculate the smooth sensitivity of the percentile selection problem using the smooth sensitivity defined by Definition \ref{def:mysmooth}. The local sensitivity remains zero until $t < 2j + 1$ and changes to one when $t \geq 2j+1$. Since the $LS_{u_p}$ is constant when $t \geq 2j+1$, the smooth sensitivity will be max when $t = 2j+1$.


\begin{proposition} 
    The smooth sensitivity (as defined in Definition \ref{def:mysmooth}) of a dataset $\db{x} \in \DB$ considering the utility function $u_p$ from Definition \ref{def:percutil} is given by
    
    $$\SSENS_{u_p, \beta}(\db{x}) = \exp(-(2j+1) \cdot \beta),$$
    where $j = \min\left(\sum_{i=0}^{k-1} u(\db{x}, x_i), \sum_{i=k+1}^{n} u(\db{x}, x_i)\right)$.
\end{proposition}

\subsection{Experimental Evaluation}
\subsubsection*{Datasets}
We tested PATENT, HEPTH, and INCOME datasets from \citeauthor{hay2016principled} \cite{hay2016principled}. The PATENT dataset contains 32,558 tuples with a high percentage of zero-valued entries at 97.80\%. In contrast, the HEPTH dataset comprises 347,414 tuples but only 21.17\% zero-valued entries, indicating more varied data. Lastly, the INCOME dataset is the largest, with 20,787,122 tuples and 44.97\% zero-valued entries, reflecting a moderate level of homogeneity in its data. An essential attribute for those datasets is the amount of $p$-th value repetitions because of the utility function.
%
%

\subsubsection*{Methods}
We consider six approaches to private percentile selection problem: i) exponential mechanism (EM) using global sensitivity; ii) permute-and-flip (PF) using global sensitivity; iii) local dampening (LD) using the element local sensitivity $\hat{\delta}(\db{x}, t, r) = LS_{u_p}(\db{x}, t)$ $= \max_{r' \in \OUT} LS_{u_p}(\db{x}, t, r)$ from utility function (Definition \ref{def:percutil}); iv) local dampening (LD2) utilizing both the utility and setup presented in \citeauthor{farias2023local}'s work, adjusted to suit our specific problem statement; v) \metname via Laplace distribution (\met-LAP) with the smooth sensitivity $\SSENS_{u_p}$; and vi) \metname via Student's T distribution (\met-T) with the smooth sensitivity $\SSENS_{u_p}$.

\subsubsection*{Evaluation}
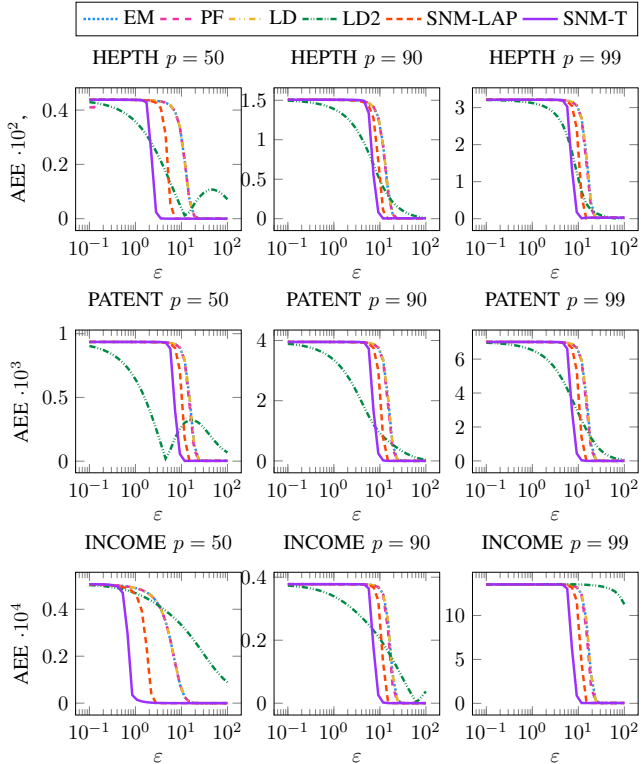
\begin{figure}[!htpb]
    \resizebox{\columnwidth}{!}{%
    \begin{tikzpicture}
        \begin{groupplot}[group style={group size= 3 by 3, vertical sep=1.7cm, horizontal sep=.56cm},width=.33\columnwidth,xlabel=$\varepsilon$,scale only axis=true,]
            \nextgroupplot[title={HEPTH $p=50$},ylabel={AEE $\cdot 10^2$,}, xmode=log, y tick label style={scaled ticks=base 10:-2}, ytick scale label code/.code={},]
            \addplot[mark=none, mark size=1.5pt, densely dotted, very thick, DodgerBlue1, y filter/.code=\pgfmathparse{abs(\pgfmathresult-41)}] table [y=err_em, x=eps]{data/final_hepth_50.dat};
            \addlegendentry{EM}
            \addplot[mark=none, mark size=1.5pt, dashed, very thick, Maroon2, y filter/.code=\pgfmathparse{abs(\pgfmathresult-41)}] table [y=err_pf, x=eps]{data/final_hepth_50.dat};
            \addlegendentry{PF}
            \addplot[mark=none, mark size=1.5pt, dashdotdotted, very thick, Goldenrod2, y filter/.code=\pgfmathparse{abs(\pgfmathresult-41)}] table [y=err_ld, x=eps]{data/final_hepth_50.dat};
            \addlegendentry{LD}
            \addplot[mark=none, mark size=1.5pt, densely dashdotdotted, very thick, SpringGreen4, y filter/.code=\pgfmathparse{abs(\pgfmathresult-41)}] table [y=err_ld2, x=eps]{data/final_hepth_50.dat};
            \addlegendentry{LD2}
            \addplot[mark=none, mark size=1.5pt, densely dashed, very thick, OrangeRed1, y filter/.code=\pgfmathparse{abs(\pgfmathresult-41)}] table [y=err_lap, x=eps]{data/final_hepth_50_lap};
            \addlegendentry{\met-LAP}
            \addplot[mark=none, mark size=1.5pt, solid, very thick, Purple1, y filter/.code=\pgfmathparse{abs(\pgfmathresult-41)}] table [y=err_t, x=eps]{data/final_hepth_50_tstudent};
            \addlegendentry{\met-T}
            \legend{};

            \nextgroupplot[title={HEPTH $p=90$}, xmode=log, y tick label style={scaled ticks=base 10:-2}, ytick scale label code/.code={},]
            \addplot[mark=none, mark size=1.5pt, densely dotted, very thick, DodgerBlue1, y filter/.code=\pgfmathparse{abs(\pgfmathresult-235.5)}] table [y=err_em, x=eps]{data/final_hepth_90.dat};
            \addlegendentry{EM}
            \addplot[mark=none, mark size=1.5pt, dashed, very thick, Maroon2, y filter/.code=\pgfmathparse{abs(\pgfmathresult-235.5)}] table [y=err_pf, x=eps]{data/final_hepth_90.dat};
            \addlegendentry{PF}
            \addplot[mark=none, mark size=1.5pt, dashdotdotted, very thick, Goldenrod2, y filter/.code=\pgfmathparse{abs(\pgfmathresult-235.5)}] table [y=err_ld, x=eps]{data/final_hepth_90.dat};
            \addlegendentry{LD}
            \addplot[mark=none, mark size=1.5pt, densely dashdotdotted, very thick, SpringGreen4, y filter/.code=\pgfmathparse{abs(\pgfmathresult-235.5)}] table [y=err_ld2, x=eps]{data/final_hepth_90.dat};
            \addlegendentry{LD2}
            \addplot[mark=none, mark size=1.5pt, densely dashed, very thick, OrangeRed1, y filter/.code=\pgfmathparse{abs(\pgfmathresult-235.5)}] table [y=err_lap, x=eps]{data/final_hepth_90_lap};
            \addlegendentry{\met-LAP}
            \addplot[mark=none, mark size=1.5pt, solid, very thick, Purple1, y filter/.code=\pgfmathparse{abs(\pgfmathresult-235.5)}] table [y=err_t, x=eps]{data/final_hepth_90_tstudent};
            \addlegendentry{\met-T}
            \legend{};

            \nextgroupplot[title={HEPTH $p=99$}, xmode=log, y tick label style={scaled ticks=base 10:-2}, ytick scale label code/.code={},]
            \addplot[mark=none, mark size=1.5pt, densely dotted, very thick, DodgerBlue1, y filter/.code=\pgfmathparse{abs(\pgfmathresult-406.15)}] table [y=err_em, x=eps]{data/final_hepth_99.dat};
            \addlegendentry{EM}
            \addplot[mark=none, mark size=1.5pt, dashed, very thick, Maroon2, y filter/.code=\pgfmathparse{abs(\pgfmathresult-406.15)}] table [y=err_pf, x=eps]{data/final_hepth_99.dat};
            \addlegendentry{PF}
            \addplot[mark=none, mark size=1.5pt, dashdotdotted, very thick, Goldenrod2, y filter/.code=\pgfmathparse{abs(\pgfmathresult-406.15)}] table [y=err_ld, x=eps]{data/final_hepth_99.dat};
            \addlegendentry{LD}
            \addplot[mark=none, mark size=1.5pt, densely dashdotdotted, very thick, SpringGreen4, y filter/.code=\pgfmathparse{abs(\pgfmathresult-406.15)}] table [y=err_ld2, x=eps]{data/final_hepth_99.dat};
            \addlegendentry{LD2}
            \addplot[mark=none, mark size=1.5pt, densely dashed, very thick, OrangeRed1, y filter/.code=\pgfmathparse{abs(\pgfmathresult-406.15)}] table [y=err_lap, x=eps]{data/final_hepth_99_lap};
            \addlegendentry{\met-LAP}
            \addplot[mark=none, mark size=1.5pt, solid, very thick, Purple1, y filter/.code=\pgfmathparse{abs(\pgfmathresult-406.15)}] table [y=err_t, x=eps]{data/final_hepth_99_tstudent};
            \addlegendentry{\met-T}
            \legend{};


            \nextgroupplot[title={PATENT $p=50$},ylabel={AEE $\cdot 10^3$}, xmode=log, y tick label style={scaled ticks=base 10:-3}, ytick scale label code/.code={},]
            \addplot[mark=none, mark size=1.5pt, densely dotted, very thick, DodgerBlue1, y filter/.code=\pgfmathparse{abs(\pgfmathresult-7758.5)}] table [y=err_em, x=eps]{data/final_patent_50.dat};
            \addlegendentry{EM}
            \addplot[mark=none, mark size=1.5pt, dashed, very thick, Maroon2, y filter/.code=\pgfmathparse{abs(\pgfmathresult-7758.5)}] table [y=err_pf, x=eps]{data/final_patent_50.dat};
            \addlegendentry{PF}
            \addplot[mark=none, mark size=1.5pt, dashdotdotted, very thick, Goldenrod2, y filter/.code=\pgfmathparse{abs(\pgfmathresult-7758.5)}] table [y=err_ld, x=eps]{data/final_patent_50.dat};
            \addlegendentry{LD}
            \addplot[mark=none, mark size=1.5pt, densely dashdotdotted, very thick, SpringGreen4, y filter/.code=\pgfmathparse{abs(\pgfmathresult-7758.5)}] table [y=err_ld2, x=eps]{data/final_patent_50.dat};
            \addlegendentry{LD2}
            \addplot[mark=none, mark size=1.5pt, densely dashed, very thick, OrangeRed1, y filter/.code=\pgfmathparse{abs(\pgfmathresult-7758.5)}] table [y=err_lap, x=eps]{data/final_patent_50_lap};
            \addlegendentry{\met-LAP}
            \addplot[mark=none, mark size=1.5pt, solid, very thick, Purple1, y filter/.code=\pgfmathparse{abs(\pgfmathresult-7758.5)}] table [y=err_t, x=eps]{data/final_patent_50_tstudent};
            \addlegendentry{\met-T}
            \legend{};

            \nextgroupplot[title={PATENT $p=90$}, xmode=log, y tick label style={scaled ticks=base 10:-3}, ytick scale label code/.code={},]
            \addplot[mark=none, mark size=1.5pt, densely dotted, very thick, DodgerBlue1, y filter/.code=\pgfmathparse{abs(\pgfmathresult-10775.0)}] table [y=err_em, x=eps]{data/final_patent_90.dat};
            \addlegendentry{EM}
            \addplot[mark=none, mark size=1.5pt, dashed, very thick, Maroon2, y filter/.code=\pgfmathparse{abs(\pgfmathresult-10775.0)}] table [y=err_pf, x=eps]{data/final_patent_90.dat};
            \addlegendentry{PF}
            \addplot[mark=none, mark size=1.5pt, dashdotdotted, very thick, Goldenrod2, y filter/.code=\pgfmathparse{abs(\pgfmathresult-10775.0)}] table [y=err_ld, x=eps]{data/final_patent_90.dat};
            \addlegendentry{LD}
            \addplot[mark=none, mark size=1.5pt, densely dashdotdotted, very thick, SpringGreen4, y filter/.code=\pgfmathparse{abs(\pgfmathresult-10775.0)}] table [y=err_ld2, x=eps]{data/final_patent_90.dat};
            \addlegendentry{LD2}
            \addplot[mark=none, mark size=1.5pt, densely dashed, very thick, OrangeRed1, y filter/.code=\pgfmathparse{abs(\pgfmathresult-10775.0)}] table [y=err_lap, x=eps]{data/final_patent_90_lap};
            \addlegendentry{\met-LAP}
            \addplot[mark=none, mark size=1.5pt, solid, very thick, Purple1, y filter/.code=\pgfmathparse{abs(\pgfmathresult-10775.0)}] table [y=err_t, x=eps]{data/final_patent_90_tstudent};
            \addlegendentry{\met-T}
            \legend{};

            \nextgroupplot[title={PATENT $p=99$}, xmode=log, y tick label style={scaled ticks=base 10:-3}, ytick scale label code/.code={},]
            \addplot[mark=none, mark size=1.5pt, densely dotted, very thick, DodgerBlue1, y filter/.code=\pgfmathparse{abs(\pgfmathresult-13835)}] table [y=err_em, x=eps]{data/final_patent_99.dat};
            \addlegendentry{EM}
            \addplot[mark=none, mark size=1.5pt, dashed, very thick, Maroon2, y filter/.code=\pgfmathparse{abs(\pgfmathresult-13835)}] table [y=err_pf, x=eps]{data/final_patent_99.dat};
            \addlegendentry{PF}
            \addplot[mark=none, mark size=1.5pt, dashdotdotted, very thick, Goldenrod2, y filter/.code=\pgfmathparse{abs(\pgfmathresult-13835)}] table [y=err_ld, x=eps]{data/final_patent_99.dat};
            \addlegendentry{LD}
            \addplot[mark=none, mark size=1.5pt, densely dashdotdotted, very thick, SpringGreen4, y filter/.code=\pgfmathparse{abs(\pgfmathresult-13835)}] table [y=err_ld2, x=eps]{data/final_patent_99.dat};
            \addlegendentry{LD2}
            \addplot[mark=none, mark size=1.5pt, densely dashed, very thick, OrangeRed1, y filter/.code=\pgfmathparse{abs(\pgfmathresult-13835)}] table [y=err_lap, x=eps]{data/final_patent_99_lap};
            \addlegendentry{\met-LAP}
            \addplot[mark=none, mark size=1.5pt, solid, very thick, Purple1, y filter/.code=\pgfmathparse{abs(\pgfmathresult-13835)}] table [y=err_t, x=eps]{data/final_patent_99_tstudent};
            \addlegendentry{\met-T}
            \legend{};


            \nextgroupplot[title={INCOME $p=50$}, xmode=log, ylabel={AEE $\cdot 10^4$}, y tick label style={scaled ticks=base 10:-4}, ytick scale label code/.code={},]
            \addplot[mark=none, mark size=1.5pt, densely dotted, very thick, DodgerBlue1, y filter/.code=\pgfmathparse{abs(\pgfmathresult-1)}] table [y=err_em, x=eps]{data/final_income_50.dat};
            \addlegendentry{EM}
            \addplot[mark=none, mark size=1.5pt, dashed, very thick, Maroon2, y filter/.code=\pgfmathparse{abs(\pgfmathresult-1)}] table [y=err_pf, x=eps]{data/final_income_50.dat};
            \addlegendentry{PF}
            \addplot[mark=none, mark size=1.5pt, dashdotdotted, very thick, Goldenrod2, y filter/.code=\pgfmathparse{abs(\pgfmathresult-1)}] table [y=err_ld, x=eps]{data/final_income_50.dat};
            \addlegendentry{LD}
            \addplot[mark=none, mark size=1.5pt, densely dashdotdotted, very thick, SpringGreen4, y filter/.code=\pgfmathparse{abs(\pgfmathresult-1)}] table [y=err_ld2, x=eps]{data/final_income_50.dat};
            \addlegendentry{LD2}
            \addplot[mark=none, mark size=1.5pt, densely dashed, very thick, OrangeRed1, y filter/.code=\pgfmathparse{abs(\pgfmathresult-1)}] table [y=err_lap, x=eps]{data/final_income_50_lap};
            \addlegendentry{\met-LAP}
            \addplot[mark=none, mark size=1.5pt, solid, very thick, Purple1, y filter/.code=\pgfmathparse{abs(\pgfmathresult-1)}] table [y=err_t, x=eps]{data/final_income_50_tstudent};
            \addlegendentry{\met-T}
            \legend{};

            \nextgroupplot[title={INCOME $p=90$}, xmode=log, y tick label style={scaled ticks=base 10:-4}, ytick scale label code/.code={},]
            \addplot[mark=none, mark size=1.5pt, densely dotted, very thick, DodgerBlue1, y filter/.code=\pgfmathparse{abs(\pgfmathresult-1296.5)}] table [y=err_em, x=eps]{data/final_income_90.dat};
            \addlegendentry{EM}
            \addplot[mark=none, mark size=1.5pt, dashed, very thick, Maroon2, y filter/.code=\pgfmathparse{abs(\pgfmathresult-1296.5)}] table [y=err_pf, x=eps]{data/final_income_90.dat};
            \addlegendentry{PF}
            \addplot[mark=none, mark size=1.5pt, dashdotdotted, very thick, Goldenrod2, y filter/.code=\pgfmathparse{abs(\pgfmathresult-1296.5)}] table [y=err_ld, x=eps]{data/final_income_90.dat};
            \addlegendentry{LD}
            \addplot[mark=none, mark size=1.5pt, densely dashdotdotted, very thick, SpringGreen4, y filter/.code=\pgfmathparse{abs(\pgfmathresult-1296.5)}] table [y=err_ld2, x=eps]{data/final_income_90.dat};
            \addlegendentry{LD2}
            \addplot[mark=none, mark size=1.5pt, densely dashed, very thick, OrangeRed1, y filter/.code=\pgfmathparse{abs(\pgfmathresult-1296.5)}] table [y=err_lap, x=eps]{data/final_income_90_lap};
            \addlegendentry{\met-LAP}
            \addplot[mark=none, mark size=1.5pt, solid, very thick, Purple1, y filter/.code=\pgfmathparse{abs(\pgfmathresult-1296.5)}] table [y=err_t, x=eps]{data/final_income_90_tstudent};
            \addlegendentry{\met-T}
            \legend{};

            \nextgroupplot[title={INCOME $p=99$}, xmode=log, y tick label style={scaled ticks=base 10:-4}, ytick scale label code/.code={},legend to name={CommonLegend},legend style={legend columns=6}]
            \addplot[mark=none, mark size=1.5pt, densely dotted, very thick, DodgerBlue1, y filter/.code=\pgfmathparse{abs(\pgfmathresult-140425)}] table [y=err_em, x=eps]{data/final_income_99.dat};
            \addlegendentry{EM}
            \addplot[mark=none, mark size=1.5pt, dashed, very thick, Maroon2, y filter/.code=\pgfmathparse{abs(\pgfmathresult-140425)}] table [y=err_pf, x=eps]{data/final_income_99.dat};
            \addlegendentry{PF}
            \addplot[mark=none, mark size=1.5pt, dashdotdotted, very thick, Goldenrod2, y filter/.code=\pgfmathparse{abs(\pgfmathresult-140425)}] table [y=err_ld, x=eps]{data/final_income_99.dat};
            \addlegendentry{LD}
            \addplot[mark=none, mark size=1.5pt, densely dashdotdotted, very thick, SpringGreen4, y filter/.code=\pgfmathparse{abs(\pgfmathresult-140425)}] table [y=err_ld2, x=eps]{data/final_income_99.dat};
            \addlegendentry{LD2}
            \addplot[mark=none, mark size=1.5pt, densely dashed, very thick, OrangeRed1, y filter/.code=\pgfmathparse{abs(\pgfmathresult-140425)}] table [y=err_lap, x=eps]{data/final_income_99_lap};
            \addlegendentry{\met-LAP}
            \addplot[mark=none, mark size=1.5pt, solid, very thick, Purple1, y filter/.code=\pgfmathparse{abs(\pgfmathresult-140425)}] table [y=err_t, x=eps]{data/final_income_99_tstudent};
            \addlegendentry{\met-T}
        \end{groupplot}

        \coordinate (c3) at ($(group c1r1)!0.5!(group c3r1)$);
        \node[above] at (c3 |- current bounding box.north) {\ref*{CommonLegend}};
    \end{tikzpicture}}
    \caption{Comparison of private selection methods for percentile selection. Plots show the absolute expected error (AEE) as a function of the privacy budget $\varepsilon \in [10^{-1}, 10^2]$. The x-axis uses log scale. Overall, \met-LAP and \met-T achieve lower expected errors than other methods for all $\varepsilon$.}
    \label{fig:res_perc}
\end{figure}

We measured the absolute expected error (AEE) of each method for every specific scenario: $|\xi(\mathscr{A}, \db{x})| = |x_k - \mathbb{E}(\mathscr{A}, \db{x})|$. Understanding each outcome's associated probability is needed to find the expected value. Meanwhile, for the exponential mechanism, permute-and-flip, and local dampening, the probabilities of each outcome are straightforward to identify through the probability mass function of each mechanism. However, finding the probability of each outcome of the \met algorithm is not straightforward.
Reasoning about the output probability of other candidates is a condition for finding the probability of the output of a particular candidate.
The specific candidate utility random variable should be greater than all others. This intricate probability function leads us to solve the following integral to find those probabilities numerically.
\begin{equation*}
    \begin{split}
        Pr[\RLEXP(\db{x}) = r] =& \int_{-\infty}^{\infty} f(i) \cdot \\ &\prod_{r \neq s} F\left(\frac{u_p(\db{x}, r) - u_p(\db{x}, s)}{N(\db{x})} + i\right) di,
    \end{split}
\end{equation*}
where $N(\db{x}) = \nicefrac{2\SSen(\db{x})}{\alpha}$, \(f\) and \(F\) represent the probability density function and the cumulative density function of the distribution used, respectively.
Figure \ref{fig:res_perc} shows the result varying the privacy budget $\varepsilon \in [10^{-1}, 10^2]$ and $p = 50, 90, 99$. For the Student's T distribution the degree of freedom was set to 3. Each dataset has a ground-truth percentile value (GT) for each percentile. The desired behavior is that with a small privacy budget, the method outputs a value near the ground-truth value.

All versions of \metname (\met-T, \met-LAP) have better accuracy when the dataset has several repetitions of the $p$-th value, i.e., a significant $j$ value. For instance, the median perspective ($p = 50$) in the PATENT dataset has $j = 0$, HEPTH has $j = 2$, and INCOME has $j = 10$. The datasets show different scenarios to assess the \met algorithms compared to the competitors. The EM and PF methods have similar expected values in all scenarios. For the HEPTH dataset with $p=50$, \met-T method achieves a similar expected value, the difference in absolute values of a maximum of 5, with $69\%$ and $85\%$ less privacy budget than LD and LD2 methods, respectively. For $p=90$, the \met-T needs less than $51\%$ and $76\%$ privacy budget than the LD and LD2 methods, respectively. For $p=99$, the behavior is similar when \met-T requires less than $51\%$ and $76\%$ budget compared with LD and LD2. For the PATENT dataset, we observe up to $51\%$, $70\%$ and $70\%$, for $p \in \{50, 90, 99\}$ respectively, in privacy budget saving when compared to LD. In the PATENT dataset with $p=50$, LD2 method quickly reaches the desired value. And for $p \in \{90, 99\}$ we observe up to $70\%$ of privacy budget saving when compared to LD2. For the INCOME dataset, when $p=50$, the difference is more evident due to the high $j$ value, but for $p \in \{90, 99\}$, the performance is quite the same in the other datasets.

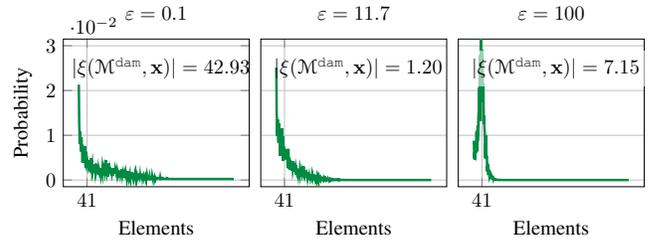
\begin{figure}
    \centering
    \resizebox{1\columnwidth}{!}{%
        \begin{tikzpicture}
            \begin{groupplot}[group style={group size= 3 by 1, horizontal sep=.2cm},width=.27\textwidth, xlabel={Elements}, xtick={41}, ymin=0, ymax=0.03,enlarge y limits=0.05, grid=both]
                \nextgroupplot[title={$\e=0.1$}, ylabel = {Probability}]
                    \addplot[mark=none, mark size=1.5pt, solid, very thick, SpringGreen4] table [y=prob, x=val]{data/ld2_behavior/hepth_50_0.1.csv};
                    \node[draw=none] at (400,0.025) {$|\xi(\MLD, \db{x})| = 42.93$};
                \nextgroupplot[title={$\e=11.7$}, ymajorticks=false]
                    \addplot[mark=none, mark size=1.5pt, solid, very thick, SpringGreen4] table [y=prob, x=val]{data/ld2_behavior/hepth_50_11.7.csv};
                    \node[draw=none] at (400,0.025) {$|\xi(\MLD, \db{x})| = 1.20$};
                \nextgroupplot[title={$\e=100$}, ymajorticks=false]
                    \addplot[mark=none, mark size=1.5pt, solid, very thick, SpringGreen4] table [y=prob, x=val]{data/ld2_behavior/hepth_50_100.0.csv};
                    \node[draw=none,fill=white, fill opacity=0.6, text opacity=1] at (400,0.025) {$|\xi(\MLD, \db{x})| = 7.15$};
            \end{groupplot}
        \end{tikzpicture}
    }
    \caption{Local Dampening (LD2) probabilities on the HEPTH dataset with \(p=50\). The first plot demonstrates that the probability of selecting element $41$ (median) is low when the privacy budget is minimal. The second graph depicts a scenario with very low expected error, suggesting that the observed low expected error occurs by chance. The last plot illustrates that with an increased privacy budget, LD2 converges effectively.} \label{fig:ld2_beh}
\end{figure}

In the LD2 experiments on the HEPTH and PATENT datasets with \( p=50 \), we observed a peculiar trend: the absolute expected error initially drops to low levels swiftly. However, as the privacy budget increases, the error, counterintuitively, increases. This rapid convergence appears to be coincidental, with the algorithm still in the process of converging. Figure \ref{fig:ld2_beh} visually captures this behavior.

%% file: sections/6-app_dt.tex
\section{Application --- Greedy Decision Tree}
Decision trees are compelling methods for classification and regression tasks \cite{kotsiantis2007supervised}. A decision tree is a graphical representation of a set of rules, where each node represents a decision based on attributes from the training dataset.

The tree topology is settled by the training algorithm that receives, as input, a dataset and outputs a decision tree. The ID3 algorithm \cite{quinlan1986induction} is one of the most known decision tree algorithms.
It recursively selects the best attribute, according to some measure, to split the data until a stopping criterion is met. In this work, the split criterion is based on the Max Operator \cite{friedman2010data}, which is the summation of each attribute value of the class with the highest frequency.
\subsection{Problem Statement}

A decision tree induction algorithm takes as input a dataset $\mathcal{T}$ with attributes $A = \{A_1, \ldots, A_d\}$ and a class attribute $C$ and produces a decision tree. The task is to build a decision tree in a differentially private manner. Specifically, we base our approach on one of the most known differentially private tree induction algorithms, the Differentially Private ID3 algorithm \cite{blum2005practical}.

\subsection{Private Mechanism and Sensitivity Analysis}
\citeauthor{blum2005practical} \cite{blum2005practical} introduced the SuLQ framework, where they design a differentially private version of ID3 as an application. The adapted application of the ID3 algorithm takes advantage of two SuLQ operators:
\begin{enumerate*}[label=\roman*)]
    \item \texttt{NoisyCount}: a Laplace mechanism operator to provide a private estimate for a count query and
    \item \texttt{Partition}: an operator that splits the dataset into disjoint subsets.
\end{enumerate*}

The primary disadvantage of the ID3 algorithm proposed by \citeauthor{blum2005practical} is its inefficient use of the privacy budget when evaluating the information gain for each attribute separately. The work presented by \citeauthor{friedman2010data} \cite{friedman2010data}, described by Algorithm \ref{alg:dpid3}, offers a more effective alternative using the exponential mechanism to evaluate each attribute independently, assessing all attributes simultaneously in a single query, resulting in the selection of an appropriate attribute for splitting. Line 13 is the exponential mechanism call that selects an attribute based on its information gain, which is the utility function.
\begin{algorithm}[htpb]
  \caption{Differentially Private ID3 (from \cite{friedman2010data}) }\label{alg:dpid3}
  \SetKwProg{Fn}{Function}{ do}{end}
  \algsetup{linenosize=\tiny}
  \footnotesize
  \Fn{\texttt{GlobalDiffPID3}( $\textrm{dataset } \mathcal{T}$, $\textrm{attribute set } A$, $\textrm{class attribute } C$,
  $\textrm{depth } d$, $\textrm{privacy budget } \varepsilon$ )} {
    $\varepsilon' \gets \nicefrac{\varepsilon}{2\cdot(d+1)}$\;
    \Return{$\texttt{BuildDiffPID3}(\mathcal{T}, A, C, d, \varepsilon')$}
  }
  \Fn{\texttt{BuildDiffPID3}( $\textrm{dataset } \mathcal{T}$, $\textrm{attribute set } A$, $\textrm{class attribute } C$,
  $\textrm{depth } d$, $\textrm{privacy budget } \varepsilon$ )} {
    $t \gets \max_{a \in A}|a|$\;
    $N_{\mathcal{T}} \gets \texttt{NoisyCount}_\varepsilon(\mathcal{T})$\;
    \If{$A = \emptyset \text{ or } d = 0 \text{ or } \nicefrac{N_{\mathcal{T}}}{t|C|} < \nicefrac{\sqrt{2}}{2}$} {
        $\mathcal{T}_c \gets \texttt{Partition}(\mathcal{T}, \forall c \in C : r_c = c)$\;
        $\forall c \in C : N_c \gets \texttt{NoisyCount}_\varepsilon(\mathcal{T}_c)$\;
        \Return{a leaf labeled with $\argmax_c N_c$}
    }
    $\bar{A} \gets \MEXP_{ig}(\mathcal{T}, \varepsilon, A)$ \tcc*[r]{Exp. mechanism call}
    $\mathcal{T}_i \gets \texttt{Partition}(\mathcal{T}, \forall i \in \bar{A} : r_{\bar{A}} = i)$\;
    $\forall i \in \bar{A} : \text{Subtree}_i \gets \texttt{BuildDiffPID3}(\mathcal{T}_i, A \backslash \bar{A}, C, d - 1, \varepsilon)$\;
    \Return{a tree with a root node labeled $\bar A$ and edges labeled 1 to $\bar A$ each going to $\text{Subtree}_i$}
  }
\end{algorithm}
The function \texttt{BuildDiffID3} in algorithm \ref{alg:dpid3} starts by checking properties like the number of attributes and the number of instances that are used as termination criteria to construct the leaves (lines 5-8). In lines 9-10, the algorithm partitions the dataset based on class labels and counts the instances for each class label. It also employs the Laplace mechanism for each class label count to select the class label for the leaf. Lines 13-16 build new decision rules recursively by privately choosing the attribute with the largest information gain value using the exponential mechanism. Moreover, it splits the dataset according to the selected attribute value and produces recursively new sub-trees for each dataset partition.

Several works address the private construction of decision trees and random forest \cite{fletcher2015differentially,fletcher2017differentially, fletcher2019decision, jagannathan2009practical, patil2014differential, rana2015differentially}. However, only \citeauthor{farias2023local} \cite{farias2023local} addresses the greedy decision tree construction algorithm applying local sensitivity. The approach proposed by \citeauthor{fletcher2017differentially} \cite{fletcher2017differentially} uses smooth sensitivity in the random forest algorithm through random decision trees. In this section, we focus on the greedy decision tree process. The following section will address the random forest application with random decision trees.

Our differentially private greedy decision tree application is similar to Algorithm \ref{alg:dpid3}. We simply replace the exponential mechanism on line 13 with our \metname, applying a utility function based on the max operator \cite{friedman2010data} that represents the summation of each attribute value of the class with the highest frequency.


\begin{definition}[Max Operator]
    Consider a dataset $\mathcal{T}$, and an attribute $A_i$, the Max operator is defined as follows:
    $
        MaxOp(\mathcal{T}, A_i) = \sum_{j \in A_i} \max_{c} \tau_{j,c}^{A_i},
    $
    where $\tau_{j,c}^{A_i}$ counts the records in $\mathcal{T}$ with attribute $A_i = j$ and class $C = c$.
\end{definition}

In our experiments, we observed that we should design a utility function representing a good split criterion and take advantage of smooth sensitivity definition to benefit from local sensitivity. Therefore, we define a utility function based on the max operator $u_{mo}$. That function outputs $1$ only for the attribute $A_i \in A$, which is the highest value of $MaxOp$ among all others $A_k \in A$, and $0$ otherwise.

\begin{definition}[Greedy decision tree utility]\label{def:gdtutil}
    Consider a dataset $\mathcal{T}$, and an attribute $A_j$, the utility is defined as:
    \[
        u_{mo}(\mathcal{T}, A_j) = \begin{cases}
            1, & \quad \text{if } A_j = \argmax\limits_{A_i \in A} MaxOp(\mathcal{T}, A_i); \\
            0, & \quad \text{otherwise}.
        \end{cases}
    \]
\end{definition}

\subsubsection*{Global Sensitivity}
The global sensitivity for $u_{mo}$ is $1$ \cite{friedman2010data}.


\subsubsection*{Local Sensitivity}
To compute the smooth sensitivity, it is crucial to have a clear understanding of the local sensitivity at a distance of $t$. Additionally, it is worth noting that the utility value will remain unchanged until $k$ additions or deletions occur in the training dataset $\mathcal{T}$. Here, $k$ refers to the difference between the highest $MaxOp$ attribute and the second-highest attribute in the dataset.


\begin{proposition}
    Let $\mathcal{T}$ be a dataset, $A_j$ an attribute, and the utility be as described in Definition \ref{def:gdtutil}. Then, the local sensitivity at distance $t$ for a greedy decision tree is:
    \[
        LS_{u_{mo}}(\mathcal{T}, t) = \begin{cases}
            1, & \quad \text{if } t \geq k; \\
            0, & \quad \text{otherwise}.
        \end{cases}
    \]
\end{proposition}

The local sensitivity remains zero until $t < k$ and changes to one when $t \geq k$. Since the $LS_{u_{mo}}$ is constant when $t < k$, the smooth sensitivity will be max when $t = k$.


\begin{proposition}
The smooth sensitivity for a greed decision tree is given by
    $
        \SSENS_{u_{mo}}(\mathcal{T}) = e^{-k \cdot \varepsilon},
    $
where $\mathcal{T}$ is a dataset, and $u_{mo}$ is the utility function described in Definition \ref{def:gdtutil}. 
\end{proposition}

\subsection{Experimental Evaluation}
\subsubsection*{Datasets}
We consider three tabular datasets:
\begin{enumerate*}[label=\roman*)]
    \item The \textit{National Long Term Care Survey (NLTCS)} \cite{manton1999national}, comprising $16$ binary attributes of $21,574$ surveyed individuals;
    \item the \textit{American Community Surveys (ACS)} dataset \cite{series2015version}, which includes information from $47,461$ rows with $23$ binary attributes, sourced from the 2013 and 2014 ACS sample sets in IPUMS-USA; and
    \item the \textit{Adult} dataset \cite{blake1998uci}, containing $45,222$ records (excluding those with missing values), featuring $12$ attributes, where $8$ are discrete and $4$ are continuous.
\end{enumerate*}

\subsubsection*{Methods}
We experimented with several mechanisms, changing the default selection algorithm described in line 13 of the Algorithm \ref{alg:dpid3}.
\begin{enumerate*}[label=\roman*)]
    \item Exponential mechanism (EM) with information gain using global sensitivity;
    \item Permute-and-flip (PF) with information gain using global sensitivity;
    \item Shifted Local dampening (SLD) with information gain using the element local sensitivity \cite{farias2023local};
    \item \metname using the Laplace Log-Normal distribution (\met-LLN);
    \item \metname using the Student's T distribution (\met-T);
    \item \metname using the Laplace distribution (\met-LAP).
\end{enumerate*}

All variants of the \metname algorithm use a utility function based on the Max Operator as the split criterion, leveraging smooth sensitivity \(\SSENS_{u_{mo}}\). Notably, \met-LLN and \met-LAP ensure approximate differential privacy (\(\delta > 0\)) rather than \(\varepsilon\)-differential privacy.
\subsubsection*{Evaluation}
\begin{figure}[htpb]
    \resizebox{\columnwidth}{!}{%
    \begin{tikzpicture}
        \begin{groupplot}[group style={group size= 2 by 3, vertical sep=1.5cm},width=.5\columnwidth,xlabel=$\varepsilon$, xmode=log]
            \nextgroupplot[title={Adult with $d=2$},ylabel={Accuracy}, ymin=0.7,scale only axis=true]
                \addplotwitherrorband{data/greedy_tree/adult_d2_em.txt}{epsilon}{acc}{std_acc}{std_acc}{DodgerBlue1}{densely dotted}{-}{EM}
                \addplotwitherrorband{data/greedy_tree/adult_d2_pf.txt}{epsilon}{acc}{std_acc}{std_acc}{Maroon2}{dashed}{x}{PF}
                \addplotwitherrorband{data/greedy_tree/adult_d2_sld.txt}{epsilon}{acc}{std_acc}{std_acc}{SpringGreen4}{dashdotdotted}{+}{SLD}
                \addplotwitherrorbanddepth{data/greedy_tree/adult_lln.txt}{epsilon}{mean}{std}{std}{OrangeRed1}{solid}{triangle}{2}{\met-LLN}
                \addplotwitherrorband{data/greedy_tree/adult_d2_tstudent.txt}{epsilon}{acc}{std_acc}{std_acc}{Purple1}{solid}{square}{\met-T}
                \addplotwitherrorbanddepth{data/greedy_tree/adult_lap.txt}{epsilon}{mean}{std}{std}{DarkOrange1}{solid}{o}{2}{\met-LAP}
                \legend{};
            \nextgroupplot[title={Adult with $d=5$}, ymin=0.66,scale only axis=true,]
                \addplotwitherrorband{data/greedy_tree/adult_d5_em.txt}{epsilon}{acc}{std_acc}{std_acc}{DodgerBlue1}{densely dotted}{-}{EM}
                \addplotwitherrorband{data/greedy_tree/adult_d5_pf.txt}{epsilon}{acc}{std_acc}{std_acc}{Maroon2}{dashed}{x}{PF}
                \addplotwitherrorband{data/greedy_tree/adult_d5_sld.txt}{epsilon}{acc}{std_acc}{std_acc}{SpringGreen4}{dashdotdotted}{+}{SLD}
                \addplotwitherrorbanddepth{data/greedy_tree/adult_lln.txt}{epsilon}{mean}{std}{std}{OrangeRed1}{solid}{triangle}{5}{\met-LLN}
                \addplotwitherrorband{data/greedy_tree/adult_d5_tstudent.txt}{epsilon}{acc}{std_acc}{std_acc}{Purple1}{solid}{square}{\met-T}
                \addplotwitherrorbanddepth{data/greedy_tree/adult_lap.txt}{epsilon}{mean}{std}{std}{DarkOrange1}{solid}{o}{5}{\met-LAP}
                \legend{};
            \nextgroupplot[title={NLTCS with $d=2$},ylabel={Accuracy},ymin=0.66,scale only axis=true,]
                \addplotwitherrorbanddepth{data/greedy_tree/nltcs_em.txt}{epsilon}{mean}{std}{std}{DodgerBlue1}{densely dotted}{-}{2}{EM}
                \addplotwitherrorbanddepth{data/greedy_tree/nltcs_pf.txt}{epsilon}{mean}{std}{std}{Maroon2}{dashed}{x}{2}{PF}
                \addplotwitherrorbanddepth{data/greedy_tree/nltcs_sld.txt}{epsilon}{mean}{std}{std}{SpringGreen4}{dashdotdotted}{+}{2}{SLD}
                \addplotwitherrorbanddepth{data/greedy_tree/nltcs_lln.txt}{epsilon}{mean}{std}{std}{OrangeRed1}{solid}{triangle}{2}{\met-LLN}
                \addplotwitherrorbanddepth{data/greedy_tree/nltcs_tstudent.txt}{epsilon}{mean}{std}{std}{Purple1}{solid}{square}{2}{\met-T}
                \addplotwitherrorbanddepth{data/greedy_tree/nltcs_lap.txt}{epsilon}{mean}{std}{std}{DarkOrange1}{solid}{o}{2}{\met-LAP}
                \legend{};
            \nextgroupplot[title={NLTCS with $d=5$},ymin=0.63,scale only axis=true,]
                \addplotwitherrorbanddepth{data/greedy_tree/nltcs_em.txt}{epsilon}{mean}{std}{std}{DodgerBlue1}{densely dotted}{-}{5}{EM}
                \addplotwitherrorbanddepth{data/greedy_tree/nltcs_pf.txt}{epsilon}{mean}{std}{std}{Maroon2}{dashed}{x}{5}{PF}
                \addplotwitherrorbanddepth{data/greedy_tree/nltcs_sld.txt}{epsilon}{mean}{std}{std}{SpringGreen4}{dashdotdotted}{+}{5}{SLD}
                \addplotwitherrorbanddepth{data/greedy_tree/nltcs_lln.txt}{epsilon}{mean}{std}{std}{OrangeRed1}{solid}{triangle}{5}{\met-LLN}
                \addplotwitherrorbanddepth{data/greedy_tree/nltcs_tstudent.txt}{epsilon}{mean}{std}{std}{Purple1}{solid}{square}{5}{\met-T}
                \addplotwitherrorbanddepth{data/greedy_tree/adult_lap.txt}{epsilon}{mean}{std}{std}{DarkOrange1}{solid}{o}{5}{\met-LAP}
                \legend{};
            \nextgroupplot[title={ACS with $d=2$},ylabel={Accuracy},ymin=0.92,scale only axis=true,]
                \addplotwitherrorbanddepth{data/greedy_tree/acs_em.txt}{epsilon}{mean}{std}{std}{DodgerBlue1}{densely dotted}{-}{2}{EM}
                \addplotwitherrorbanddepth{data/greedy_tree/acs_pf.txt}{epsilon}{mean}{std}{std}{Maroon2}{dashed}{x}{2}{PF}
                \addplotwitherrorbanddepth{data/greedy_tree/acs_sld.txt}{epsilon}{mean}{std}{std}{SpringGreen4}{dashdotdotted}{+}{2}{SLD}
                \addplotwitherrorbanddepth{data/greedy_tree/acs_lln.txt}{epsilon}{mean}{std}{std}{OrangeRed1}{solid}{triangle}{2}{\met-LLN}
                \addplotwitherrorbanddepth{data/greedy_tree/acs_tstudent.txt}{epsilon}{mean}{std}{std}{Purple1}{solid}{square}{2}{\met-T}
                \addplotwitherrorbanddepth{data/greedy_tree/acs_lap.txt}{epsilon}{mean}{std}{std}{DarkOrange1}{solid}{o}{2}{\met-LAP}
                \legend{};
            \nextgroupplot[title={ACS with $d=5$}, ymin=0.85,legend to name={CommonLegend1},legend style={legend columns=3},scale only axis=true,]
                \addplotwitherrorbanddepth{data/greedy_tree/acs_em.txt}{epsilon}{mean}{std}{std}{DodgerBlue1}{densely dotted}{-}{5}{EM}
                \addplotwitherrorbanddepth{data/greedy_tree/acs_pf.txt}{epsilon}{mean}{std}{std}{Maroon2}{dashed}{x}{5}{PF}
                \addplotwitherrorbanddepth{data/greedy_tree/acs_sld.txt}{epsilon}{mean}{std}{std}{SpringGreen4}{dashdotdotted}{+}{5}{SLD}
                \addplotwitherrorbanddepth{data/greedy_tree/acs_lln.txt}{epsilon}{mean}{std}{std}{OrangeRed1}{solid}{triangle}{5}{\met-LLN}
                \addplotwitherrorbanddepth{data/greedy_tree/acs_tstudent.txt}{epsilon}{mean}{std}{std}{Purple1}{solid}{square}{5}{\met-T}
                \addplotwitherrorbanddepth{data/greedy_tree/acs_lap.txt}{epsilon}{mean}{std}{std}{DarkOrange1}{solid}{o}{5}{\met-LAP}

        \end{groupplot}

        \coordinate (c3) at ($(group c1r1)!0.5!(group c3r1)$);
        \node[above] at (c3 |- current bounding box.north) {\ref*{CommonLegend1}};
    \end{tikzpicture}}
    \caption{Comparison of private selection methods for the greedy decision tree application. The plots show the mean accuracy of greedy decision tree experiments - 5 runs of 10-fold cross-validation, where $d \in \{2, 5\}$ and $\varepsilon \in \{0.01, 0.05, 0.1, 0.5, 1, 2\}$. X axis is in log scale. All \met variants consistently achieve superior accuracy compared to competing methods. Notably, the performance of \met-T is especially significant, as it ensures $\varepsilon$-dp.}
    \label{fig:res_gdt}
\end{figure}
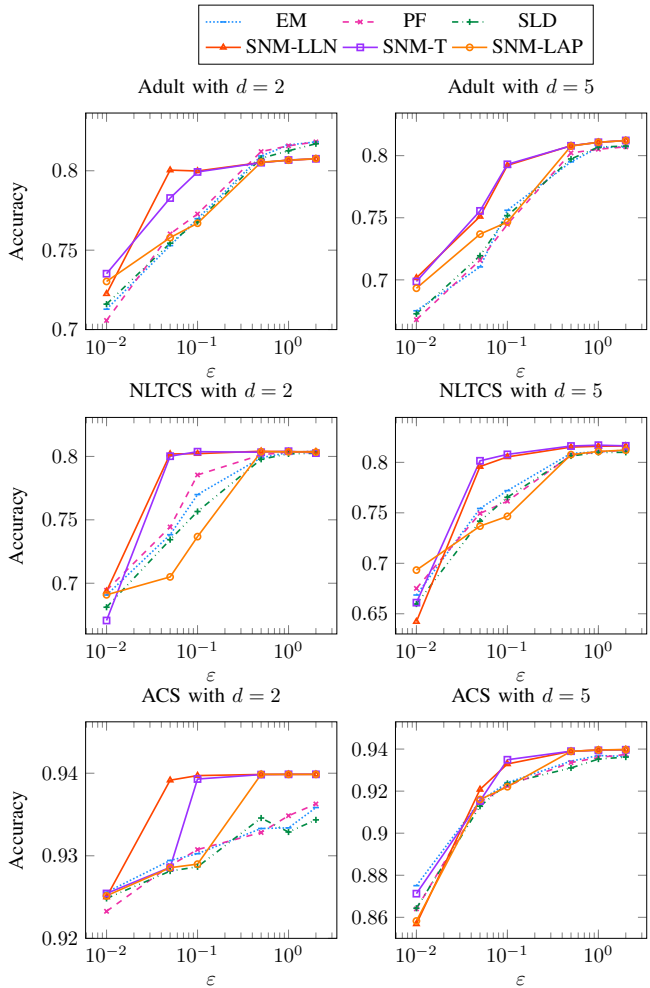

We measured the accuracy of each mechanism varying the max tree depth $d \in \{2,5\}$ and the privacy budget $\varepsilon \in \{0.01, 0.05, 0.1, 0.5, 1.0, 2.0\}$. Each trial was measured using $10$-fold validation, and each scenario ran $5$ times. Figure \ref{fig:res_gdt} shows the average accuracy of those scenarios. \looseness=-1

We observed that \met with Student's T and Laplace Log-Normal are the best-performing method in most scenarios. In the Adult dataset with $d=2$, \met-LLN and \met-T are the best-performing methods for small $\varepsilon$. However, when the budget is higher, all the competitors have better accuracy, indicating that with the Adult dataset with shallow trees ($d=2$), the information gain split criteria works better than the max operator even with a higher signal-to-sensitivity ratio when compared with the max operator. Nevertheless, with $d=5$, \met-LLN and \met-T perform better for all $\varepsilon$ values. When the dataset is NLTCS, \met-T has up to $8.58\%$ of improvement in accuracy when compared with the competitors. \met-T improves up to $1.15\%$ with the ACS dataset compared to the other methods.

%% file: sections/7-app_rf.tex
\section{Application --- Random Forest}
\label{chap:apps}
Classification based on decision tree algorithms are remarkable tools for data mining \cite{friedman2010data}.
They also serve as core building block for random forests \cite{breiman2001random}.
Random forest is a supervised learning algorithm that combines the predictions of several decision trees, an ensemble of predictors. The algorithm starts by building a set of decision trees and then applies a majority voting to the outcomes of those trees.

The decision tree is a supervised learning algorithm based on a tree structure, where each intermediate node represents a decision based on a feature, and each leaf node represents a label. The algorithm starts from the root node and, based on comparing the feature value with a threshold on numerical features, it splits the tree.
If the feature value exceeds the threshold, the algorithm goes to the right child node. Otherwise, it goes to the left child node. When the feature selected is categorical, the node has one child for each possible categorical value, and the comparison is made by checking the equality of attribute value. The algorithm continues until it reaches a leaf node when the node's majority label is the tree's outcome.

This section presents an application of a differentially private random forest algorithm using the \metname as a selection mechanism. The method is a random decision tree designed to save privacy budget in the splitting process. We describe and test the random forest algorithm with several selection mechanisms, including our \metname, under different scenarios and datasets to compare its results against our competitors.

\subsection{Problem Statement}
A random forest algorithm takes as input a dataset $\db{x}$ with attributes $F = \{F_1, \ldots, F_d\}$, a max depth parameter $h$, and a parameter $c$ that represents the forest size. The task is to build a forest with $c$ trees $\mathcal{T} = \{\tau_1, \ldots, \tau_c \}$ in a differential private manner.

\subsection{Random Decision Trees}
The most common approaches to building a decision tree are ID3 \cite{quinlan1986induction}, CART \cite{breiman1984classification}, and C4.5 \cite{quinlan2014c4}. They are based on some purity measures as splitting criteria. However, they have a lower generalization performance \cite{breiman2001random}. To overcome this problem, the random decision tree algorithm applies random splitting criteria. The generalization helps ensemble methods like the random forest to add diversity to the ensemble and, therefore, improve the performance \cite{fletcher2019decision}.

In a greedy decision tree algorithm, the splitting process of a node depends on the input data in the same way that the leaf node class counts dictated by the data, which may leak some information. Considering information leakage, we should prevent these privacy breaches using differential privacy. We must spend some privacy budget whenever data needs to be queried. So, seeking to save privacy budget, the random decision trees apply random split criteria to avoid the usage of privacy budget and save it for the leaf node class counts queries \cite{fletcher2019decision}.

\citeauthor{fletcher2017differentially} \cite{fletcher2017differentially} propose a random forest algorithm based on random decision trees that satisfies differential privacy. The algorithm applies the exponential mechanism in the leaves to select the majority label.

The work of \citeauthor{fletcher2017differentially} is summarized in Algorithm \ref{alg:buildtree}, which starts by splitting the dataset into $c$ chunks. Then, each chunk $x_i$ builds a random tree $\tau_i$. Finally, the algorithm applies the exponential mechanism to select the majority label of the forest and adds the tree to the forest $\mathcal{T}$. \looseness=-1

\begin{algorithm}
  \caption{Random Forest Algorithm \cite{fletcher2017differentially} }\label{alg:buildtree}
  \algsetup{linenosize=\tiny}
  \footnotesize
  \SetKwProg{Fn}{Function}{ do}{end}
  \Fn{\texttt{buildForest}($\text{Dataset } \db{x}$, $\text{Forest Size } c$, $\text{Features } F$, $\text{Depth } h$)} {
    \For{$i \in \texttt{split}({\db{x}, c}) $} {
      $\tau \gets \texttt{setMajority}(\texttt{buildTree}({x_i, F, h, 0}))$\;
      $\mathcal{T} \gets \mathcal{T} \cup \tau$ \;
    }
  }
  \Fn{\texttt{buildTree}($\text{Dataset } \db{x}$, $\text{Features } F$, $\text{Max Depth } h$, $\text{Depth } d$)} {
    $T \gets \{\}$\;
    \If{$d < h$} {
      Uniformly select attribute $f$ from $F$ to split current node\;
      \If{$f$ is continuous} {
        Uniformly select split point $p$ from the $f$'s domain\;
        $\db{x}_{l}, \db{x}_{r} \gets \texttt{split}(\db{x}, f, p)$ \;
        $T \cup \texttt{buildTree}(\db{x}_{l}, F, h, d+1) \cup \texttt{buildTree}(\db{x}_{r}, F, h, d+1)$\;
      }
    } \Else {
      $F \gets F \backslash f$ \;
      \ForAll{$a \in f$} {
        $\db{x}_a \gets \texttt{getData}(\db{x}, f, a)$ \;
        $T \cup \texttt{buildTree}(\db{x}_{a}, F, h, d+1)$\;
      }
    }
    \Return{$T$}
  }
\end{algorithm}

In line 2, the dataset is partitioned into $c$ chunks and iterated over it. The build tree function is called in line 3. The build tree function is a conventional recursive approach in that the features are randomly chosen for each node and the split point using only the data's domains, regardless of the data itself. The novel part of the proposed algorithm is the set majority function also in line 3. The set majority function applies the exponential mechanism to select the majority label of the leaf node through a specifically designed utility function. The proposed utility function, shown by Definition \ref{def:utilitytree}, outputs $1$ for the label with the highest count in the leaf node and $0$ otherwise.

\begin{definition}[Utility Function \cite{fletcher2017differentially}] \label{def:utilitytree}
    The utility function $u$ is defined as:
    \begin{equation} \label{eq:utilitytree}
        u(\db{x}, r) = \begin{cases}
            1, & \quad \text{if } r = \argmax_{i \in \OUT} n_i; \\
            0, & \quad \text{otherwise}.
        \end{cases}
    \end{equation}
    where $n_i$ is the number of samples of class $i$ in the leaf node. \looseness=-1
\end{definition}

The global sensitivity of the utility function (Definition \ref{def:utilitytree}) is $1$. The work of \citeauthor{fletcher2017differentially} \cite{fletcher2017differentially} applies the smooth sensitivity instead of global sensitivity to reach a better signal-to-noise ratio. However, as proven in the Theorem \ref{the:sam} below, it does not satisfy differential privacy.

\begin{theorem} \label{the:sam}
	The exponential mechanism setting $\MEXP_{\score, \varepsilon}(\db{x}, r) \propto \exp\left({\frac{\varepsilon \score(\db{x}, r)}{2 \SSen(\db{x})}} \right)$ does not satisfy $\varepsilon$-differential privacy with smooth sensitivity instead of global sensitivity.

    \begin{proof}
	Assuming that the exponential mechanism with smooth sensitivity satisfies $\varepsilon$-differential privacy, consider an approval voting example. Here, voters can endorse multiple candidates instead of choosing just one. In this scenario, the utility function assigns a value of $1$ to the candidate with the highest votes and $0$ to all others.

    The utility function exhibits a smooth sensitivity of \(\SSen(\db{x}) = \exp\left(-j \varepsilon\right)\), where \(j\) is the vote disparity between the top candidate and the runner-up in dataset \(\db{x}\) (Theorem \ref{the:smoothutilitytree}). The local sensitivity of the utility function \(u\) remains zero until the vote gap \(j\) is large enough to affect the comparison, at which point it jumps to 1. The smooth sensitivity peaks when \(t = j\), yielding \(\SSen(\db{x}) = \exp\left(-j \varepsilon\right)\).

    For example, consider the output set $\OUT = [\text{C1}, \text{C2}, \text{C3}, \text{C4}, \text{C5}]$ with the vote count vector $\mathbf{v} = [22, 8, 17, 4, 0]$ from dataset $\db{x}$. Here, candidate C1 leads with 22 votes, followed by others, with C2 receiving 8 votes, and so forth. The utility function (Definition \ref{def:utilitytree}) assigns a score of $1$ solely to candidate C1. The vote difference between the leading candidate and the second-most voted, denoted as $j$, is 5.

    To ensure the differential privacy definition is necessary to address all possible neighboring datasets from $\db{x}$, for instance, the dataset $\db{y}$ by adding one more vote for the second-most voted candidate (C3). Therefore, the $j$ parameter reduces to the value for $4$, implying a smooth sensitivity value of $0.135$. Using the privacy budget as $0.5$, we have $Pr[\MEXP_{\score, 0.5}(\db{x}, \text{C3})] = 0.04$ and $Pr[\MEXP_{\score, 0.5}(\db{y}, \text{C3})] = 0.10$, following the Definition \ref{def:dp}:
	  \begin{align*}
		  Pr[\MEXP_{\score, 0.5}(\db{x}, \text{C3})] &\leq e^{0.5} Pr[\MEXP_{\score, 0.5}(\db{y}, \text{C3})] \Rightarrow \\
		  0.04 &\leq 0.16 \Rightarrow \top\\
		  Pr[\MEXP_{\score, 0.5}(\db{y}, \text{C3})] &\leq e^{0.5} Pr[\MEXP_{\score, 0.5}(\db{x}, \text{C3})] \Rightarrow\\
		  0.10 &\leq 0.07 \Rightarrow \bot
	  \end{align*}

    Therefore, by contradiction, the exponential mechanism setting does not hold $\varepsilon$-differential privacy with smooth sensitivity instead of global sensitivity.
	\end{proof}
\end{theorem}

To address that issue, we replace the exponential mechanism with our \met in
\citeauthor{fletcher2017differentially}'s random forest algorithm as the differentially private selection procedure.

\begin{algorithm}[htbp]
  \algsetup{linenosize=\tiny}
  \footnotesize
  \caption{Set Majority Labels with \metname}\label{alg:setmajor}
  \SetKwProg{Fn}{Function}{ do}{end}
  \Fn{\texttt{setMajority}($\text{Tree}~\tau$)} {
    \For{$l \in \ell$} {
      $labelCounts \gets l.counts$ \;
      $l.maj \gets \texttt{\met}({labelCounts})$\;
    }
  }
\end{algorithm}

Algorithm \ref{alg:setmajor} details our set majority function, which implements the \metname algorithm. It begins by traversing all leaves of the tree $\tau$ (line 2). For each leaf $l$, the algorithm retrieves the label counts of the leaf node in line 3. Subsequently, the \metname algorithm is applied to select the majority label of the leaf node in line 4. It is important to note that to execute the \metname algorithm, the smooth sensitivity of the utility function is required, as demonstrated in Theorem \ref{the:smoothutilitytree}.

\begin{theorem}[Smooth sensitivity of Def. \ref{def:utilitytree} \cite{fletcher2019decision}] \label{the:smoothutilitytree}
    The smooth sensitivity of the utility function $u$ (definition \ref{def:utilitytree}) is:
    $\SSen(\db{x}) = \exp\left({-j \varepsilon} \right)$,
    where $j$ is the difference between the most frequent and the second-most frequent labels in $\db{x}$.
\end{theorem}

\subsection{Experimental Evaluation}
This section presents the datasets, methods, and experimental evaluation results. We selected six datasets to evaluate the performance of our proposed method compared with other baselines.

\subsubsection*{Methods}
Our evaluation employs the standard random forest algorithm (Algorithm \ref{alg:buildtree}). We term the non-private implementation of Algorithm \ref{alg:buildtree} as WDP. The experiment employs various selection mechanisms, including the exponential mechanism (EM), permute-and-flip (PF), local dampening mechanism (LD), \metname with Laplace Log-Normal distribution (\met-LLN), \metname with Student's T distribution (\met-T), and \metname with Laplace distribution (\met-LAP). We configure all privacy-preserving mechanisms, excluding the non-private method, with the utility function defined in Definition \ref{def:utilitytree}. EM and PF utilize a global sensitivity of $1.0$. We empirically determine the element local sensitivity across each dataset for local dampening. We follow Theorem \ref{the:smoothutilitytree} to find the smooth sensitivity within our \metname. Our goal is to measure the accuracy impact of choosing the \metname as a private selection method.

\begin{figure}[!bth]
  \resizebox{\columnwidth}{!}{%
  \begin{tikzpicture}
      \begin{groupplot}[group style={group size= 2 by 3, vertical sep=1.5cm},width=.5\columnwidth,xlabel=$\varepsilon$, xmode=log,scale only axis=true,]
        \nextgroupplot[title={Adult},ylabel={Accuracy}, xmode=log]
        \addplot[dashed, thick] coordinates { (0.01,0.77488) (2,0.77488) };
        \addlegendentry{WDP}
        \addplot[mark=-, densely dotted, thick, mark size=1.5pt, DodgerBlue1] table [y=mean_acc, x=budget]{figures/adult_em.dat};
        \addlegendentry{EM}
        \addplot[mark=x, dashed, thick, mark size=1.5pt, Maroon2] table [y=mean_acc, x=budget]{figures/adult_pf.dat};
        \addlegendentry{PF}
        \addplot[mark=+, dashdotdotted, thick, mark size=1.5pt, SpringGreen4] table [y=mean_acc, x=budget]{figures/adult_ld.dat};
        \addlegendentry{LD}
        \addplot[mark=triangle, solid, thick, mark size=1.5pt, OrangeRed1] table [y=mean_acc, x=budget]{figures/adult_lln.txt};
        \addlegendentry{\met-LLN}
        \addplot[mark=square, solid, thick, mark size=1.5pt, Purple1] table [y=mean_acc, x=budget]{figures/adult_tstudent.txt};
        \addlegendentry{\met-T}
        \addplot[mark=o, solid, thick, mark size=1.5pt, DarkOrange1] table [y=mean_acc, x=budget]{figures/adult_lap.txt};
        \addlegendentry{\met-LAP}
        \legend{};

        \nextgroupplot[title={Mushroom}, xmode=log]
        \addplot[dashed, semithick, mark size=1.5pt] coordinates { (0.01,0.792246) (2,0.792246) };
        \addlegendentry{WDP}
        \addplot[mark=-, densely dotted, thick, mark size=1.5pt, DodgerBlue1] table [y=mean_acc, x=budget]{figures/mushroom_em.dat};
        \addlegendentry{EM}
        \addplot[mark=x, dashed, thick, mark size=1.5pt, Maroon2] table [y=mean_acc, x=budget]{figures/mushroom_pf.dat};
        \addlegendentry{PF}
        \addplot[mark=+, dashdotdotted, thick, mark size=1.5pt, SpringGreen4] table [y=mean_acc, x=budget]{figures/mushroom_ld.dat};
        \addlegendentry{LD}
        \addplot[mark=triangle, solid, thick, mark size=1.5pt, OrangeRed1] table [y=mean_acc, x=budget]{figures/mushroom_lln.txt};
        \addlegendentry{\met-LLN}
        \addplot[mark=square, solid, thick, mark size=1.5pt, Purple1] table [y=mean_acc, x=budget]{figures/mushroom_tstudent.txt};
        \addlegendentry{\met-T}
        \addplot[mark=o, solid, thick, mark size=1.5pt, DarkOrange1] table [y=mean_acc, x=budget]{figures/mushroom_lap.txt};
        \addlegendentry{\met-LAP}
        \legend{};

        \nextgroupplot[title={Wine},xmode=log]
        \addplot[dashed, semithick, mark size=1.5pt] coordinates { (0.01,0.4591836734693877) (2,0.4591836734693877) };
        \addlegendentry{WDP}
        \addplot[mark=-, densely dotted, thick, mark size=1.5pt, DodgerBlue1] table [y=mean_acc, x=budget]{figures/wine_em.dat};
        \addlegendentry{EM}
        \addplot[mark=x, dashed, thick, mark size=1.5pt, Maroon2] table [y=mean_acc, x=budget]{figures/wine_pf.dat};
        \addlegendentry{PF}
        \addplot[mark=+, dashdotdotted, thick, mark size=1.5pt, SpringGreen4] table [y=mean_acc, x=budget]{figures/wine_ld.dat};
        \addlegendentry{LD}
        \addplot[mark=triangle, solid, thick, mark size=1.5pt, OrangeRed1] table [y=mean_acc, x=budget]{figures/wine_lln.txt};
        \addlegendentry{\met-LLN}
        \addplot[mark=square, solid, thick, mark size=1.5pt, Purple1] table [y=mean_acc, x=budget]{figures/wine_tstudent.txt};
        \addlegendentry{\met-T}
        \addplot[mark=o, solid, thick, mark size=1.5pt, DarkOrange1] table [y=mean_acc, x=budget]{figures/wine_lap.txt};
        \addlegendentry{\met-LAP}
        \legend{};

        \nextgroupplot[title={Pen-digits},xmode=log]
        \addplot[dashed, semithick, mark size=1.5pt] coordinates { (0.01,0.7259663483401546) (2,0.7259663483401546) };
        \addlegendentry{WDP}
        \addplot[mark=-, densely dotted, thick, mark size=1.5pt, DodgerBlue1] table [y=mean_acc, x=budget]{figures/pen_em.dat};
        \addlegendentry{EM}
        \addplot[mark=x, dashed, thick, mark size=1.5pt, Maroon2] table [y=mean_acc, x=budget]{figures/pen_pf.dat};
        \addlegendentry{PF}
        \addplot[mark=+, dashdotdotted, thick, mark size=1.5pt, SpringGreen4] table [y=mean_acc, x=budget]{figures/pen_ld.dat};
        \addlegendentry{LD}
        \addplot[mark=triangle, solid, thick, mark size=1.5pt, OrangeRed1] table [y=mean_acc, x=budget]{figures/pen_lln.txt};
        \addlegendentry{\met-LLN}
        \addplot[mark=square, solid, thick, mark size=1.5pt, Purple1] table [y=mean_acc, x=budget]{figures/pen_tstudent.txt};
        \addlegendentry{\met-T}
        \addplot[mark=o, solid, thick, mark size=1.5pt, DarkOrange1] table [y=mean_acc, x=budget]{figures/pen_lap.txt};
        \addlegendentry{\met-LAP}
        \legend{};

        \nextgroupplot[title={Compas},xmode=log]
        \addplot[dashed, semithick, mark size=1.5pt] coordinates { (0.01,0.2570221752903907) (2,0.2570221752903907) };
        \addlegendentry{WDP}
        \addplot[mark=-, densely dotted, thick, mark size=1.5pt, DodgerBlue1] table [y=mean_acc, x=budget]{figures/compas_em.dat};
        \addlegendentry{EM}
        \addplot[mark=x, dashed, thick, mark size=1.5pt, Maroon2] table [y=mean_acc, x=budget]{figures/compas_pf.dat};
        \addlegendentry{PF}
        \addplot[mark=+, dashdotdotted, thick, mark size=1.5pt, SpringGreen4] table [y=mean_acc, x=budget]{figures/compas_ld.dat};
        \addlegendentry{LD}
        \addplot[mark=triangle, solid, thick, mark size=1.5pt, OrangeRed1] table [y=mean_acc, x=budget]{figures/compas_lln.txt};
        \addlegendentry{\met-LLN}
        \addplot[mark=square, solid, thick, mark size=1.5pt, Purple1] table [y=mean_acc, x=budget]{figures/compas_tstudent.txt};
        \addlegendentry{\met-T}
        \addplot[mark=o, solid, thick, mark size=1.5pt, DarkOrange1] table [y=mean_acc, x=budget]{figures/compas_lap.txt};
        \addlegendentry{\met-LAP}
        \legend{};

        \nextgroupplot[title={Wall-sensor}, xmode=log,legend to name={CommonLegend2},legend style={legend columns=4}]
        \addplot[dashed, semithick, mark size=1.5pt] coordinates { (0.01,0.6521978021978022) (2,0.6521978021978022) };
        \addlegendentry{WDP}
        \addplot[mark=-, densely dotted, thick, mark size=1.5pt, DodgerBlue1] table [y=mean_acc, x=budget]{figures/wall_em.dat};
        \addlegendentry{EM}
        \addplot[mark=x, dashed, thick, mark size=1.5pt, Maroon2] table [y=mean_acc, x=budget]{figures/wall_pf.dat};
        \addlegendentry{PF}
        \addplot[mark=+, dashdotdotted, thick, mark size=1.5pt, SpringGreen4] table [y=mean_acc, x=budget]{figures/wall_ld.dat};
        \addlegendentry{LD}
        \addplot[mark=triangle, solid, thick, mark size=1.5pt, OrangeRed1] table [y=mean_acc, x=budget]{figures/wall_lln.txt};
        \addlegendentry{\met-LLN}
        \addplot[mark=square, solid, thick, mark size=1.5pt, Purple1] table [y=mean_acc, x=budget]{figures/wall_tstudent.txt};
        \addlegendentry{\met-T}
        \addplot[mark=o, solid, thick, mark size=1.5pt, DarkOrange1] table [y=mean_acc, x=budget]{figures/wall_lap.txt};
        \addlegendentry{\met-LAP}

      \end{groupplot}

      \coordinate (c3) at ($(group c1r1)!0.5!(group c2r1)$);
      \node[above] at (c3 |- current bounding box.north) {\ref*{CommonLegend2}};
  \end{tikzpicture}}
  \caption{Comparison of private selection methods for the random forest problem. The plots show mean accuracy for WDP, EM, PF, LD, and \met variants of random forest with 32 random trees varying $\varepsilon \in \{0.01, 0.05, 0.1, 1, 2\}$. X is in log scale. The \met flavors constantly reach the standard non-private random forest accuracy level. When compared with other private selection methods, the variants of \met surpass in almost all $\varepsilon$ values.}
  \label{fig:res_rf}
\end{figure}
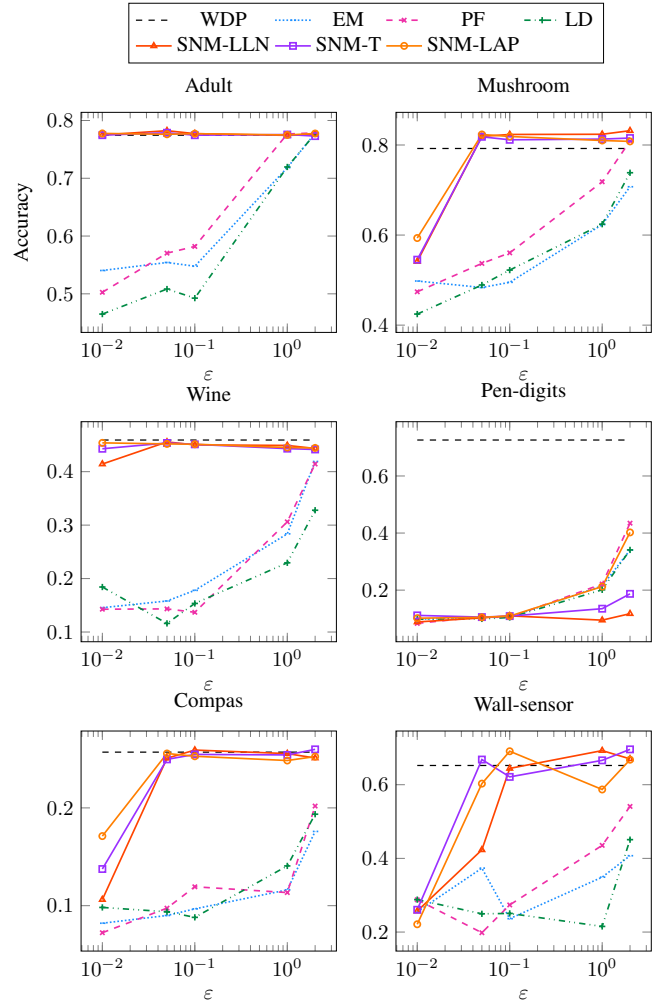

\subsubsection*{Evaluation}
We measured the accuracy of those methods over ten executions using the accuracy metric. The process split the dataset in $80\%$ for the training step and $20\%$ for evaluation purposes. The privacy budget varies by $\{0.01, 0.05, 0.1, 1, 2\}$. We also set each random forest with 32 trees. The max tree depth was set for each dataset using the Theorem 2 from \citeauthor{fletcher2019decision}'s work \cite{fletcher2019decision}.

\subsubsection*{Results}
Our experimental procedure compares our method using the accuracy metric in 6 datasets. The datasets were selected based on their size, number of features, and number of classes. The Adult dataset comprises 48,842 instances with 6 continuous and 8 discrete features, a maximum tree depth of 9, and 2 classes. Compas contains 4,732 entries, 9 continuous and 4 discrete features, a depth of 5, and 11 classes. The Wine dataset involves 4,898 samples, all 11 features being continuous, a depth of 10, and 7 classes. Mushroom includes 8,124 entries, 22 discrete features, a maximum depth of 11, and 2 classes. The Pen-digits dataset, one of the largest with 109,092 instances, features 17 continuous attributes, a depth of 12, and 10 classes. Finally, Wall-sensor offers 5,456 samples, 4 continuous features, a depth of 4, and 4 classes. Figure \ref{fig:res_rf} shows the result of our proposed random forest algorithm using the \metname with the other selection algorithms varying the budget parameter. \looseness=-1


Firstly, we focus on the experiments using the Mushroom \cite{misc_mushroom_73} and Adult \cite{misc_adult_2} datasets. Both datasets have mostly discrete attributes and few classes but differ in size. The Adult dataset has more than 48 thousand tuples compared to almost 8 thousand in the Mushroom dataset. The max depth was set to 9 and 11 for Adult and Mushroom datasets. Looking at the results of our experiment in the Adult dataset, we can observe that even with a small privacy budget, we can deliver excellent accuracy results, achieving the version without private guarantees. Using the Mushroom dataset, when the budget is $0.1$, all the versions of \metname surpasses the standard random forest, i.e., our private method is better than the privateless version. By our method, the randomness input can improve the power of the tree generalization \cite{breiman2001random}, leading to better accuracy. Fixing with a privacy budget of $1$, our method is similar to permute-and-flip's accuracy performance, but using the privacy budget of $0.01$, we can deliver the same accuracy performance as the permute-and-flip with $100$ times more budget (see Figure \ref{fig:res_rf}). \looseness=-1

The Wine Quality dataset \cite{cortez2009modeling} has almost 5 thousand records with 11 continuous features and zero discrete features. The Pen-Based Recognition of Handwritten Digits (Pen-digits) dataset \cite{misc_pen_81} has more than 109 thousand records with 17 continuous features and zero discrete features. The maximum depth was set to 10 and 12 for the wine and pen-digit datasets. The random forest results with all the versions of \met reach almost the non-private version in the wine dataset, outperforming all the adversaries even with very low privacy budget values. In the experiments using the pen-digits dataset, all the private methods underperform, mainly because of the dataset's $j$ (Theorem \ref{the:smoothutilitytree}) value. The difference between the highest class count and the second highest is narrow in the pen-digits dataset, implying a smooth sensitivity almost equal to global sensitivity.

The Compas and the Wall-sensor datasets have similar sizes but differ in the number of classes. The Compas (Correctional Offender Management Profiling for Alternative Sanctions) dataset \cite{githubGitHubPropublicacompasanalysis} has 11 classes, and the Wall-Following Robot Navigation Dataset (Wall-sensor) \cite{misc_wall_data_194} has only four classes. Figure \ref{fig:res_rf} shows that our proposed \met versions outperform the private selection adversaries using the Compas and wall-sensor datasets. Even with many classes, the proposed random forest algorithm employing \met with a small privacy budget reaches the standard random forest without any privacy concerns.

%% file: sections/8-conclusion.tex
\section{Conclusion}
This paper introduces the \metname, a novel differentially private selection algorithm. We formally describe our approach, its privacy attributes, and its utility. We demonstrate that under mild conditions, our algorithm's utility, leveraging the Laplace distribution, consistently matches or exceeds that of competing methods while satisfying differential privacy criteria. \metname utilizes local sensitivity across various private selection scenarios. Additionally, we comprehensively compare our mechanism against established methods such as local dampening, report-noisy-max, permute-and-flip, and exponential mechanisms.
We empirically evaluated our approach on three different applications:
\begin{enumerate*}[label=\roman*)]
    \item Percentile selection;
    \item Greedy decision trees; and,
    \item Random forest.
\end{enumerate*}

In the experiments, we faced a limitation of our proposed algorithm. The notion of local sensitivity at a distance $t$ quickly converges to global sensitivity due to the max operator (from Definition \ref{def:lst}) iterating over all the possible outcomes. It was necessary to design specific utility functions to overcome that situation, e.g., only the best answer has a non-zero utility. Another limitation of smooth sensitivity is its computational complexity, which leads to algorithms with high time demands. To address this challenge, we have employed simple utility functions that the smooth sensitivity is analytically solvable.

We presume that applying the notion of element local sensitivity \cite{farias2023local} should solve this limitation. As future work, we aim to prove the use of the element local sensitivity with the \metname. Additionally, formalizing problems as single functions can be challenging, as many real-world situations are complex and involve multiple objectives. While theoretically possible, combining these through a single function, such as weighting or ranking them, might not always perfectly capture the nuances of some problems. Therefore, differentially private multi-objective selection is a research topic in our pipeline.